\documentclass[]{bytedance_seed}

\usepackage[toc,page,header]{appendix}

\titleformat*{\paragraph}{\normalsize\sffamily\bfseries}

\renewcommand{\emph}[1]{\textit{#1}}
\usepackage{wrapfig}
\usepackage{url}
\usepackage{nicefrac}
\usepackage{minitoc}
\usepackage{enumitem}
\usepackage{amsfonts}
\usepackage{amssymb}
\usepackage{xspace}
\usepackage{paralist}
\usepackage{array}
\usepackage{makecell}
\usepackage{wrapfig}
\usepackage[utf8]{inputenc}
\usepackage{booktabs}
\usepackage{multirow}
\usepackage{pifont}
\usepackage{tabularx}
\usepackage{adjustbox}
\setcitestyle{square,numbers,comma}
\usepackage{float}

\newcolumntype{C}[1]{>{\centering\arraybackslash}m{#1}}
\newcolumntype{L}[1]{>{\raggedright\arraybackslash}m{#1}}
\definecolor{posgreen}{RGB}{0,128,64}
\definecolor{negred}{RGB}{180,50,50}
\newcommand{\posdiff}[1]{{\color{posgreen}{+#1}}}
\newcommand{\negdiff}[1]{{\color{negred}{-#1}}}
\definecolor{MidnightBlue}{RGB}{25,25,112}

\newcommand{\modelname}{\textsc{\textbf{MMProLong}}\xspace}
\newcommand{\extractsingle}{\texttt{extract-single}\xspace}
\newcommand{\extractmulti}{\texttt{extract-multi}\xspace}
\newcommand{\reasoningtask}{\texttt{reasoning}\xspace}
\newcommand{\ocrfull}{\texttt{OCR-full}\xspace}
\newcommand{\ocrneedle}{\texttt{OCR-needle}\xspace}
\newcommand{\promptplaceholder}[1]{\colorbox{gray!12}{\textcolor{MidnightBlue}{\ttfamily #1}}}

\title{Training Long-Context Vision-Language Models Effectively with Generalization Beyond 128K Context}

\author[1,2, *]{Zhaowei~Wang}
\author[2]{Lishu~Luo}
\author[2]{Haodong~Duan}
\author[2]{Weiwei~Liu}
\author[2]{Sijin~Wu}
\author[2]{Ji~Luo}
\author[2]{Shen~Yan}
\author[2]{Shuai~Peng}
\author[2]{Sihang~Yuan}
\author[2]{Chaoyi~Huang}
\author[2, \dagger]{Yi~Lin}
\author[1, \dagger]{Yangqiu~Song}

\affiliation[1]{CSE Department, HKUST}
\affiliation[2]{ByteDance Seed}

\contribution[*]{Work done at ByteDance Seed}
\contribution[\dagger]{Corresponding authors}

\abstract{
Long-context modeling is becoming a core capability of modern large vision-language models (LVLMs), enabling sustained context management across long-document understanding, video analysis, and multi-turn tool use in long-horizon agents. 
Yet practical training recipes remain insufficiently explored, 
particularly for designing and balancing long-context data mixtures.
In this work, we present a systematic study of \emph{long-context continued pre-training} (LongPT) for LVLMs, extending a 7B model from 32K to 128K context with extensive ablations on long-document data.
We first show that long-document VQA is substantially more effective than OCR transcription.
Building on this observation, our ablations further yield three key findings:
\begin{inparaenum}[i)]
\item for sequence-length distribution, balanced data outperforms target-length-focused data (e.g., 128K), suggesting that long-context ability requires generalizable key-information retrieval across various lengths and positions;
\item retrieval remains the primary bottleneck, favoring retrieval-heavy mixtures with modest reasoning data for task diversity;
\item pure long-document VQA largely preserves short-context capabilities, suggesting that instruction-formatted long data reduces the need for short-data mixing.
\end{inparaenum}
Based on these findings, we introduce \modelname, obtained by long-context continued pre-training from Qwen2.5-VL-7B with only a 5B-token budget. 
\modelname improves long-document VQA scores by 7.11\% and maintains strong performance at 256K and 512K contexts beyond its 128K training window, without additional training.
It further generalizes to webpage-based multimodal needle retrieval, long-context vision-text compression, and long-video understanding without task-specific supervision.
Overall, our study establishes a practical LongPT recipe and an empirical foundation for advancing long-context vision-language models.
}

\checkdata[Email]{\texttt{\{zwanggy, yqsong\}@cse.ust.hk};\ \texttt{linyi.james@bytedance.com}}

\checkdata[Code]{\url{https://github.com/zhaowei-wang-nlp/MMProLong}}

\begin{document}
\maketitle

%不需要目录就注释掉 注意目录不要和第一页放在一块 要有\newpage
%\newpage
%\tableofcontents
%\newpage

% !TEX root = ../paper.tex

\section{Introduction}
\label{sec:introduction} 
The ability to process long context has unlocked a wide range of new capabilities for both large language models~\citep[LLMs;][]{yang2025qwen3,llama4} and large vision-language models~\citep[LVLMs;][]{bai2025qwen3,seed2_0}. 
For LVLMs in particular, long-context modeling enables multi-hop reasoning over document collections~\citep{wang2025mmlongbench,wang2024needle}, capturing spatiotemporal dependencies from hour-long videos~\citep{wu2024longvideobench,fu2025video}, and maintaining context consistency in long-horizon agent tasks~\citep{geng2025webwatcher,zhang2026browsecomp,zhang2026clawbench}.

To support such capabilities, LVLMs' context windows have been rapidly scaled to 128K tokens and beyond, driven by both proprietary models (e.g., Gemini 3.1~\citep{gemini_3_1_pro_2026} and GPT-5.4~\citep{openai2026introducing_gpt5_4}) and open-weight alternatives such as Qwen3-VL~\citep{bai2025qwen3} and GLM-4.5V~\citep{hong2025glm}. 
However, recent technical reports~\citep{bai2025qwen3,hong2025glm} provide only limited details on the use of long-document data, leaving practical recipes for developing long-context vision-language models (LCVLMs) insufficiently explored. 
It remains unclear which types of long-context data to synthesize, how to mix different long-context tasks and incorporate short-context data, and how training choices such as length distributions affect the resulting model.

To bridge this gap, we present a systematic study of the \emph{long-context continued pre-training} (LongPT) in LVLMs. 
Building on Qwen2.5-VL-7B~\citep{bai2025qwen2}, we extend its context window from 32K to 128K and study how to construct and combine multimodal long-context training data.
We use long documents as data sources because they provide realistic multimodal contexts with complex visual layouts and dense textual content.
From these documents, we construct five training tasks grouped into two task categories: long-document VQA and OCR transcription.
Comparing these tasks, we find that long-document VQA is substantially more effective than OCR transcription, suggesting that instruction-formatted supervision and task diversity ranging from information extraction to complex numerical reasoning are important for LongPT.

Having established long-document VQA as the primary data source, we then study practical training designs for LongPT in LVLMs, covering sequence-length distribution, long-context task mixtures, and the role of short-context data. 
In these ablations, we observe three main findings: 
\begin{inparaenum}[i)]
\item for sequence-length distribution, we find balanced data outperforms target-length-focused data near 128K, suggesting that LongPT should teach generalizable key-information retrieval across various lengths and positions rather than specialize to a single target length;
\item key-information retrieval remains the primary bottleneck in long-context pre-training, favoring retrieval-heavy mixtures with modest reasoning data to maintain task diversity; and
\item unlike LLM long-context pre-training practice~\citep{gao2025prolong}, pure long-document VQA largely preserves short-context capabilities, suggesting that instruction-formatted long data reduces the need for short-context mixing.
\end{inparaenum}

In light of these observations, we arrive at the final LongPT recipe and train our model, \modelname, with a 5B-token budget.
It improves long-document VQA performance by 7.11\% at 64K and 128K contexts and maintains strong performance at 256K and 512K without additional training or adaptation, exceeding baselines by over 20\%.
These gains also transfer to broader multimodal long-context tasks, including webpage-based needle-in-a-haystack on MM-NIAH~\citep{wang2024needle}, long-context compression on VTCBench~\citep{zhao2025vtcbench}, and long-video understanding~\citep{fu2025video,zhou2025mlvu,wu2024longvideobench}.
Finally, we validate the recipe on Qwen3-VL~\citep{bai2025qwen3}, showing that it is not specific to Qwen2.5-VL and can benefit stronger long-context backbones.
Together, these results suggest a practical path toward training long-context vision-language models with data-efficient, transferable LongPT recipes.
\section{Related Work}
\label{sec:related}
\noindent\textbf{Context window extension.}
Extending the context window has become a key direction for improving long-context performance, with recent LLMs supporting 128K and even 1M context windows~\citep{yang2025qwen3,hurst2024gpt,team2024gemini_1_5,google2024gemini_2,google2025gemini2_5}.
Existing approaches either extend context windows through lightweight methods, such as positional extrapolation~\citep{chen2023extending,pengyarn,ding2024longrope,zhang2024extending,zhu2023pose} and attention modifications~\citep{chen2023longlora,xiao2023efficient,xiao2024infllm,bertsch2023unlimiformer,jin2024llm}, or rely on continued pre-training to build more robust long-context capability~\citep{xiong2024effective,fu2024data,gao2025prolong,Yang2025Qwen251MTR}.
Our work follows the continued pre-training method, but studies it in multimodal settings where long contexts contain interleaved image and text tokens.

\noindent\textbf{Long-context vision-language models.}
As LLM context windows have expanded, recent LVLMs such as Gemini 3.1 Pro~\citep{gemini_3_1_pro_2026}, Claude Opus 4.7~\citep{anthropic2026introducing_claude_opus_4_7}, and Qwen3-VL~\citep{bai2025qwen3} have also supported substantially longer contexts.
However, recent LVLM technical reports~\citep{bai2025qwen3,hong2025glm} reveal limited details about how long-context capability is actually built, leaving practical LongPT recipes underexplored.
Concurrent work~\citep{veselka2026longvlm} studies long-document data construction for LVLMs, but mainly builds on backbones that already support 128K or longer contexts, such as Qwen3-VL~\citep{bai2025qwen3} and Mistral 3.1~\citep{mistralai2025mistral_small_3_1}; thus, its findings may reflect context alignment rather than true context extension.
For example, they find that 1B-token LongPT outperforms its 10B-token counterpart, and that LongSFT outperforms LongPT.
In contrast, we study LongPT on Qwen2.5-VL~\citep{bai2025qwen2}, whose native context window is only 32K, allowing us to directly examine how to extend LVLMs to longer context.
Another line of work studies long-video understanding~\citep{chen2024longvila,shen2025longvita,yang2025eea,zhang2024longva,yang2025longvt,liu2025bolt,shen2024longvu}, but these methods are often specialized for temporal redundancy and video token reduction rather than general long-context LVLM training.

\noindent\textbf{Multimodal long-context evaluation.}
Recent benchmarks evaluate multimodal long-context understanding from diverse perspectives, including long-document VQA~\citep{ma2024mmlongbenchdoc,deng2025longdocurl}, multimodal needle-in-a-haystack~\citep{wang2024needle,wu2024visual}, vision-text compression~\citep{zhao2025vtcbench}, and long-video understanding~\citep{fu2025video,zhou2025mlvu,wang2025lvbench}.
Among them, MMLongBench~\citep{wang2025mmlongbench} provides a comprehensive evaluation across five task categories with standardized context lengths up to 128K.
Our evaluation covers MMLongBench, VTCBench, and long-video benchmarks, demonstrating the broad generalization of our model \modelname.
\section{Experimental Setup}
\label{sec:setup}
We conduct our LongPT experiments using Qwen2.5-VL-7B~\citep{bai2025qwen2}, extending its original 32K context window to 128K.
Following the Dynamic-NTK heuristic~\citep{dynamicntk}, we scale the mRoPE base frequency from its original value of $1\times10^6$ to $4\times10^6$ with detailed ablations provided in \cref{app:complementary_experiment_results:mrope_base}.
Each LongPT run is trained with a fixed budget of 5B tokens, a maximum sequence length of 131,072 tokens, and a global batch size of 4M tokens.
Throughout the paper, we use binary prefixes: $K=2^{10}$, $M=2^{20}$, and $B=2^{30}$.
We provide the full implementation details in \cref{app:final_recipe:implementation_details} and the full evaluation details in \cref{app:evaluation_details}.
\section{Multimodal Long-Context Data Curation}
\label{sec:curation}
Recent studies~\citep{team2025gemma3,google2026gemma4,hong2025glm} have identified data synthesis and mixture design as critical factors in pre-training, making data design a central focus of our study.
For LVLMs, documents provide a natural source for synthesizing image-text data, as each page combines rich visual layout with dense textual content and can be rendered into long multimodal sequences.
In this section, we first describe the preliminary step for constructing the document pool, which provides the raw image-text source for further data synthesis.
Next, we discuss five training tasks for synthesizing multimodal long-context data from long documents, grouped into two categories: long-document VQA and OCR transcription.
Finally, we conduct experiments to evaluate which task category provides more effective LongPT supervision.

\subsection{Document Pool Construction}
\label{sec:curation:document_pool_construction}
To support scalable data synthesis, we first construct a large-scale document pool comprising over 1.5 million PDF-formatted documents from multiple sources. 
The resulting pool spans a broad range of document types, including academic papers, books, and technical manuals, as well as diverse domains such as engineering, medicine, social sciences, and biology. 
Detailed statistics and domain distribution are provided in \cref{app:document_pool_stats}.
For data synthesis, we select documents with 32 to 50 pages from this pool.
With the 2$\times$2-pixel unshuffle in our Qwen2.5-VL backbone, these documents yield multimodal sequences ranging from 32K to 128K tokens. 
To avoid evaluation contamination, we further filter out potential overlap with evaluation benchmarks using SHA-256 hashes of PDF content.

As LVLMs operate on images rather than PDF files, each PDF page is rendered to an image at $DPI=144$ using PyMuPDF~\footnote{https://github.com/pymupdf/pymupdf}.
This resolution provides a practical trade-off between visual fidelity and storage cost.
In addition, we use an OCR expert model fine-tuned from Seed~2.0~\citep{seed2_0} to parse each rendered page into layout-aware blocks.
These parsed blocks are further used in both task categories: title and section labels provide the section structure to guide the sampling of coherent page segments for long-document VQA, while recognized text blocks serve as transcription targets for the OCR transcription training tasks.

\subsection{Long-Document VQA Data Synthesis}
\label{sec:curation:long_doc_vqa_method}

\noindent\textbf{Segment-level synthesis pipeline.}
We construct the long-document VQA training data using a short-to-long synthesis pipeline.
The key idea is to generate a QA pair from a short, semantically coherent page segment, and then place it back into the full-document context to form a long-context training instance.

Specifically, we first parse each document with our OCR expert model and identify its section structure using two element labels, namely \texttt{title} and \texttt{section}.
Based on the parsed structure, we randomly sample one or more consecutive sections whose total length spans 8--15 pages.
This produces a coherent page segment at the section level for QA generation.

Next, we feed the page images of the sampled segment into Seed~2.0~\citep{seed2_0}, which serves as the QA-generator.
We prompt the model to generate a QA pair, along with evidence descriptions and evidence pages, using the detailed prompt provided in \cref{app:long_document_vqa_details:prompt}.

Finally, we recover the original full document corresponding to the sampled segment and combine it with the generated QA pair.
This yields a single long-context VQA training instance, where the answer can be inferred from a localized short segment while the model must process the full long-document context.

\begin{wrapfigure}{r}{0.48\textwidth}
    \vspace{-13pt}
    \centering
    \includegraphics[width=0.47\textwidth]{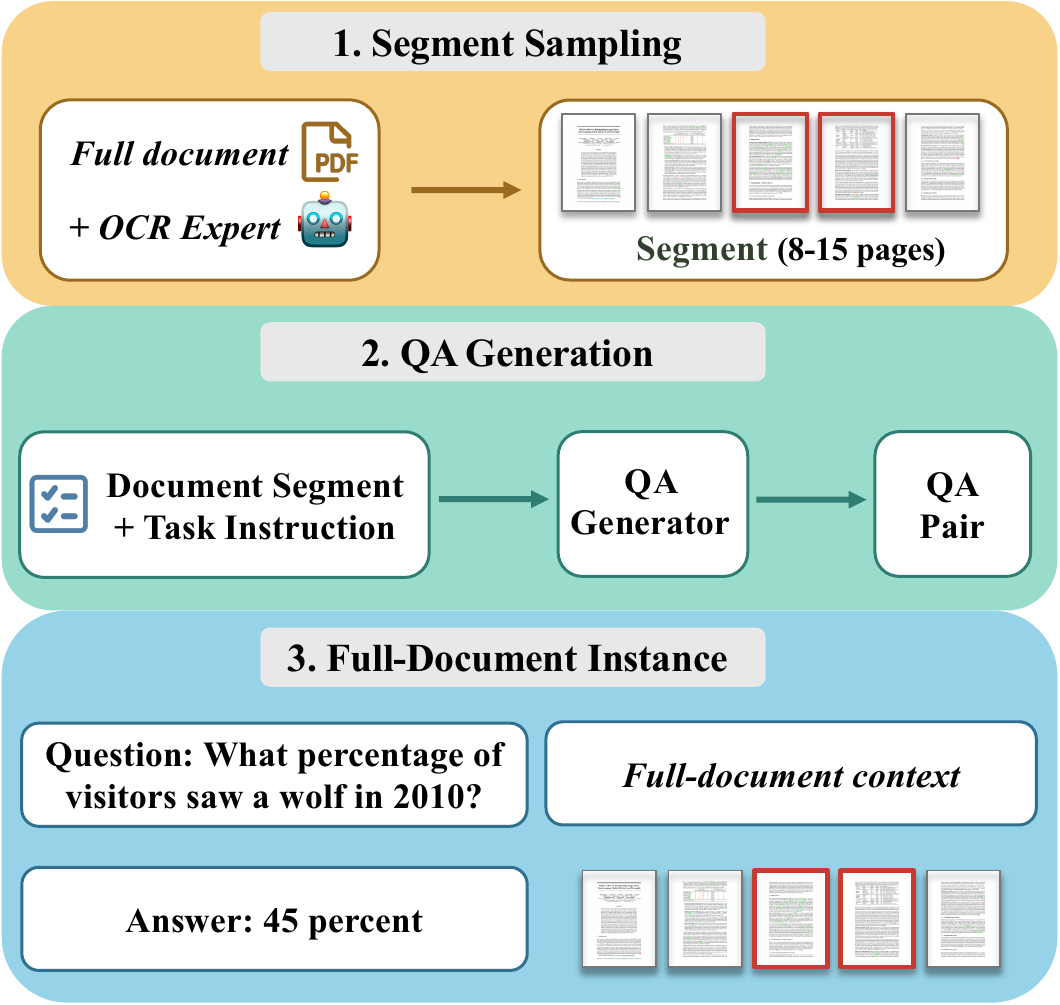}
    \caption{\textbf{Long-document VQA synthesis pipeline.} We first parse the full document with an OCR expert and sample a coherent segment. An LVLM used as a QA generator (e.g., Seed 2.0) then produces a QA pair from the segment, which is inserted back into the original document to form a long-context training instance.}
    \label{fig:long_vqa_synthesis_method}
    \vspace{-15pt}
\end{wrapfigure}

\noindent\textbf{Data quality and efficiency.}
Since we provide the QA-generation model with only 8--15 pages, the pipeline relies on strong short-context understanding, without requiring full-document processing.
In this way, we find that the generated QA pairs are of high quality, and further verify them through a manual check described in \cref{app:long_document_vqa_details:verification}.
By sampling short segments, this pipeline is also efficient, substantially reducing the cost of generating large-scale data.

A key challenge in segment-level QA synthesis is ensuring that locally valid questions remain unambiguous when evaluated in the full-document context.
Specifically, because QA pairs are generated from a short segment, the same question may have a different answer when placed back into the full document.
For example, a question such as \textit{``What is the reported revenue?''} may be answerable within the sampled section, but ambiguous in a full financial report where different sections report revenue for different departments or years.
To avoid such global-context false positives, we require the QA-generation model to add explicit segment anchors to the question, such as \textit{``in the Introduction section''} or \textit{``on pages~20--25''}.

\noindent\textbf{Data types.}
With this segment-level QA synthesis pipeline, we synthesize three training tasks of long-document VQA data, each targeting a distinct capability defined by the type and number of evidence pieces required to answer the question.
They cover increasing evidence complexity:
\begin{inparaenum}[(i)]
\item \textbf{single-page extraction} (\extractsingle) asks the model to retrieve factual information from a single page, e.g., ``According to the Homemade Bitters recipe on Page 39, how long should the herbs soak in vodka?'';
\item \textbf{multi-page extraction} (\extractmulti) requires the model to aggregate factual information from multiple pages, e.g., ``Based on Pages 6, 13, and 19, list all risk factors mentioned in the report.''; and
\item \textbf{reasoning} (\reasoningtask) further requires numerical or logical operations over extracted information, such as summation, comparison, or counting across pages, e.g., ``What is the difference between total consumption and total imports for rice production in 2020?''.
\end{inparaenum}

Together, the first two training tasks focus on locating and extracting relevant evidence from long documents, while the reasoning task further evaluates whether the model can operate on the extracted evidence.

\subsection{OCR Transcription Data Synthesis}
\label{sec:curation:ocr_transcription_method}
In addition to long-document VQA, another category of long-context training tasks we build is OCR transcription.
This task category encourages LVLMs to capture long-distance image-text dependencies by requiring them to transcribe text elements across all pages of a long document.

\noindent\textbf{Synthesis pipeline.}
For each document, we first parse every page with our OCR expert and retain text elements such as section titles, paragraphs, tables, and captions.
We then construct an OCR transcription sequence by using the rendered page images as visual input and the parsed text elements as the target output.
With this formulation, LVLMs must repeatedly attend back to the dense textual content in the rendered images and transcribe it over long distances, thereby modeling long-distance image-text dependencies.

\noindent\textbf{Data types.}
Using this pipeline, we generate two training tasks of OCR transcription data.
These types are defined by the scope of pages to be transcribed:  \begin{inparaenum}[(i)]
\item \textbf{full-document OCR} (\ocrfull) requires the model to transcribe text elements from all pages of the document, encouraging dense image-text dependency modeling across the full context; and
\item \textbf{needle-page OCR} (\ocrneedle) selects only a small subset of pages (1--3 pages) for transcription and keeps the remaining pages as distractors, turning OCR transcription into a retrieval-style long-context training task.
\end{inparaenum}
Collectively, these two tasks encourage LVLMs to model long-distance image-text dependencies under both dense transcription and retrieval-style settings.

\subsection{Comparing Long-Document VQA and OCR Transcription}
\label{sec:curation:data_method_compare}
We compare the five candidate tasks under a controlled 5B-token budget. For each task, we build a separate training set and train Qwen2.5-VL-7B~\cite{bai2025qwen2} using the hyperparameters in \cref{sec:setup}. Dataset statistics, such as token counts and sequence-length distributions, are provided in \cref{app:long_document_vqa_details:data_stats,app:ocr_transcription_details:data_stats}.

\noindent\textbf{SFT after LongPT on OCR Transcription.}
OCR transcription encourages long-distance image-text dependency modeling but is not naturally aligned with instruction-following evaluations.
To favor OCR-based LongPT, we further apply a 5B-token SFT stage to the OCR-trained checkpoints using LLaVA-OneVision instruction data~\citep{li2024llava}; data details are in \cref{app:short_context_data_details}.

\begin{table*}[t]
    \centering
    \caption{We compare long-document VQA data with OCR transcription data under the same setting. The models are evaluated on the document category of MMLongBench~\citep{wang2025mmlongbench} at 64K and 128K, which contains three datasets: MMLongBench-Doc~\citep{ma2024mmlongbenchdoc}, LongDocURL~\citep{deng2025longdocurl}, and SlideVQA~\citep{tanaka2023slidevqa}. We abbreviate them as MMLB-D, LD-URL, and SLIDE, respectively. \textbf{SFT} means an extra 5B-token SFT stage.}
    \label{tab:vqa_effectiveness}
    \setlength{\tabcolsep}{4pt}
    \resizebox{0.9\textwidth}{!}{%
    \begin{tabular}{l*{4}{C{1.2cm}}|*{4}{C{1.2cm}}|C{1.8cm}}
    \toprule
    & \multicolumn{4}{c|}{\textbf{64K MMLongBench}} & \multicolumn{4}{c|}{\textbf{128K MMLongBench}} & \multirow{2}{*}{\textbf{AVG.}} \\
     \cmidrule(lr){2-5} \cmidrule(lr){6-9}
    \textbf{Training data} & \multicolumn{1}{c}{\footnotesize MMLB-D} & \multicolumn{1}{c}{\footnotesize LD-URL} & \multicolumn{1}{c}{\footnotesize SLIDE} & \multicolumn{1}{c|}{\footnotesize AVG.} & \multicolumn{1}{c}{\footnotesize MMLB-D} & \multicolumn{1}{c}{\footnotesize LD-URL} & \multicolumn{1}{c}{\footnotesize SLIDE} & \multicolumn{1}{c|}{\footnotesize AVG.} & \\
    \midrule
    Qwen2.5-VL-7B & 32.17 & 49.57 & 75.00 & 52.24 & 26.96 & 51.85 & 68.00 & 48.94 & 50.59\,{\scriptsize\makebox[2em][l]{}} \\
    \midrule
    \extractsingle & 33.85 & 59.73 & 77.00 & 56.86 & 30.89 & 55.69 & 77.00 & 54.53 & 55.69\,{\scriptsize\makebox[2em][l]{\posdiff{5.1}}} \\
    \extractmulti & 32.75 & \textbf{64.32} & 77.00 & \textbf{58.02} & \textbf{31.50} & 54.82 & \textbf{81.00} & \textbf{55.77} & \textbf{56.90}\,{\scriptsize\makebox[2em][l]{\posdiff{6.3}}} \\
    \reasoningtask & 32.67 & 60.34 & \textbf{79.00} & 57.33 & 29.23 & \textbf{61.61} & 76.00 & 55.62 & 56.47\,{\scriptsize\makebox[2em][l]{\posdiff{5.9}}} \\
    \midrule
    \ocrfull & 23.67 & 11.06 & 59.00 & 31.24 & 23.97 & 20.36 & 61.00 & 35.11 & 33.17\,{\scriptsize\makebox[2em][l]{\negdiff{17.4}}} \\
    \ocrneedle & \textbf{40.07} & 28.77 & 68.00 & 45.61 & 22.75 & 37.24 & 66.00 & 42.00 & 43.80\,{\scriptsize\makebox[2em][l]{\negdiff{6.8}}} \\
    \ocrfull (SFT) & 37.19 & 53.07 & 78.00 & 56.09 & 25.03 & 51.73 & 78.00 & 51.59 & 53.84\,{\scriptsize\makebox[2em][l]{\posdiff{3.2}}} \\
    \ocrneedle (SFT) & 33.79 & 50.38 & 78.00 & 54.06 & 26.65 & 49.84 & 76.00 & 50.83 & 52.44\,{\scriptsize\makebox[2em][l]{\posdiff{1.9}}} \\
    \bottomrule
    \end{tabular}
    }

\end{table*}

\noindent\textbf{Long-document VQA provides stronger supervision.}
The results are shown in \cref{tab:vqa_effectiveness}. 
First, the 32K base model degrades substantially at 128K, with MMLongBench-Doc dropping from 32.17\% to 26.96\%.
More importantly, OCR transcription tasks yield poor downstream performance, especially full-document OCR, whose overall average drops by 17.4\% to 33.17\%.
After adding the SFT stage to improve instruction-following ability, the OCR-trained checkpoints obtain moderate gains of 3.24\% and 1.85\% for full-document and needle-page OCR, respectively.
In contrast, all three long-document VQA tasks consistently improve performance by more than 5\% in absolute terms, with multi-page extraction achieving the best average of 56.90\%.
This makes long-document VQA a stronger and more computationally efficient supervision source for LongPT, yielding better downstream performance without an additional 5B-token SFT stage. 
Its advantage suggests that instruction-formatted supervision and task diversity, ranging from information extraction to complex numerical reasoning, are important for LongPT.
We therefore focus on long-document VQA in the remaining data-design experiments.

\section{Data Mixture and Training Design}
\label{sec:mixing}
Having identified long-document VQA as an effective data source, we now study how to turn it into a practical LongPT recipe.
Specifically, we examine three key design choices: the distribution of training instance lengths, the mixture of long-context data, and the preservation of short-context performance.
We provide an additional ablation on the RoPE base frequency in \cref{app:complementary_experiment_results:mrope_base}.

\begin{table*}[t]
    \centering
    \caption{Long-context data mixture test. We grid search the mix of information extraction vs.\ reasoning tasks from $0:10$ to $10:0$ under the fixed 5B-token budget. \textbf{Ratio} represents the extraction-to-reasoning ratio.}
    \label{tab:extract_reason_ratio}
    \setlength{\tabcolsep}{4pt}
    \resizebox{0.85\textwidth}{!}{%
    \begin{tabular}{l*{4}{C{1.2cm}}|*{4}{C{1.2cm}}|C{1.2cm}}
    \toprule
    & \multicolumn{4}{c|}{\textbf{64K MMLongBench}} & \multicolumn{4}{c|}{\textbf{128K MMLongBench}} & \multirow{2}{*}{\textbf{AVG.}} \\
    \cmidrule(lr){2-5} \cmidrule(lr){6-9}
    \textbf{Ratio} & \multicolumn{1}{c}{\footnotesize MMLB-D} & \multicolumn{1}{c}{\footnotesize LD-URL} & \multicolumn{1}{c}{\footnotesize SLIDE} & \multicolumn{1}{c|}{\footnotesize AVG.} & \multicolumn{1}{c}{\footnotesize MMLB-D} & \multicolumn{1}{c}{\footnotesize LD-URL} & \multicolumn{1}{c}{\footnotesize SLIDE} & \multicolumn{1}{c|}{\footnotesize AVG.} & \\
    \midrule
    0:10 & 32.67 & 60.34 & 79.00 & 57.33 & 29.23 & \textbf{61.61} & 76.00 & 55.62 & 56.47 \\
    2:8 & 32.04 & \textbf{65.02} & 77.00 & 58.02 & 27.37 & 61.34 & 74.00 & 54.24 & 56.13 \\
    4:6 & 31.57 & 59.49 & 78.00 & 56.35 & 32.95 & 53.39 & \textbf{79.00} & 55.11 & 55.73 \\
    6:4 & 32.52 & 64.86 & 79.00 & 58.79 & 32.72 & 55.53 & \textbf{79.00} & 55.75 & 57.27 \\
    8:2 & \textbf{36.00} & 62.69 & \textbf{80.00} & \textbf{59.56} & \textbf{34.19} & 56.33 & 77.00 & 55.84 & \textbf{57.70} \\
    10:0 & 33.98 & 59.48 & 79.00 & 57.49 & 32.12 & 59.07 & 78.00 & \textbf{56.40} & 56.94 \\
    \bottomrule
    \end{tabular}
    }
\end{table*}

\subsection{Training Sequence-Length Distribution}
\label{sec:mixing:length}
\begin{wrapfigure}{r}{0.45\textwidth}
    \vspace{-25pt}
    \centering
    \includegraphics[width=0.40\textwidth]{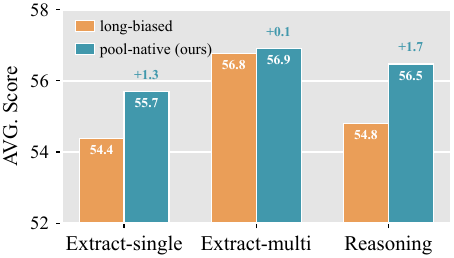}
    \caption{Comparison between different length distributions. We report overall average scores across 64K and 128K of the long-document VQA from MMLongBench. See full results in \cref{app:complementary_experiment_results:length_distribution}.}
    \label{fig:length_distribution_avg}
    \vspace{-10pt}
\end{wrapfigure}
When extending the context window of LLMs, prior work~\citep{fu2024data,gao2025prolong} often relies on books or code repositories from SlimPajama~\citep{soboleva2023slimpajama} or the Stack~\citep{kocetkov2022stack}, whose sequence lengths are naturally distributed across 8K to 128K tokens.
In contrast, our long-document pool contains a large number of documents ranging from 20 to 200 pages, providing sufficient coverage for constructing training instances at different target lengths.
This raises a practical question: how should we choose the length distribution of synthesized training instances?

\noindent\textbf{Constructing data with different length distributions.}
Here, we study two length distributions, namely \emph{pool-native} and \emph{long-biased}.
In the data-curation study (\cref{sec:curation}), we use the \emph{pool-native} length distribution by default, as training instances are synthesized from documents naturally sampled within the 32--50 page range, without additional length-based reweighting.

Given that we evaluate our models at a 128K context length, it is natural to ask whether allocating the token budget to longer examples leads to better LongPT results.
We therefore construct the \emph{long-biased} variant of data in which 83.9\% of the examples contain at least 100K tokens, compared with only 23.6\% in the \emph{pool-native} distribution (See \cref{tab:over_100k_ratio}).
This variant exposes the model more frequently to near-maximum-length contexts (128K), whereas the pool-native distribution covers a broader range of context lengths.
Detailed statistics for both distributions are summarized in \cref{app:long_document_vqa_details:data_stats,app:long_document_vqa_details:data_stats_long_only}.

\noindent\textbf{A diverse length distribution yields better long-context capability.}
The average performance of both length distributions is summarized in \cref{fig:length_distribution_avg}, with full evaluation results provided in \cref{app:complementary_experiment_results:length_distribution}.
Overall, the \emph{pool-native} distribution outperforms the \emph{long-biased} distribution, yielding average gains of $+1.3$, $+0.1$, and $+1.7$ points for \extractsingle, \extractmulti, and \reasoningtask tasks, respectively.
Notably, the \emph{pool-native} distribution consistently matches or outperforms the \emph{long-biased} distribution while containing fewer near-128K training examples.

These empirical findings suggest that long-context ability is not a discrete capability acquired only at a specific target length, such as 128K context.
Instead, it requires continuous calibration across different absolute positions and relative image-text distances.
In other words, LongPT should teach the model to retrieve key information in a way that generalizes across diverse long-context scenarios.
Based on this observation, we adopt the \emph{pool-native} distribution for all the following experiments in this paper.

\begin{table*}[t]
    \centering
    \caption{\textbf{Short-context performance under different short-data mixing ratios.}
    We report the average over six short-context benchmarks together with per-benchmark scores. \textbf{Short Data} means the proportion of short-context data. $0\%$ means only using long-context data.}
    \label{tab:short_mix_short_performance}
    \setlength{\tabcolsep}{4pt}
    \resizebox{0.85\textwidth}{!}{%
    \begin{tabular}{l||*{2}{C{1.2cm}}|*{3}{C{1.2cm}}|C{1.2cm}|C{1.2cm}}
    \toprule
    & \multicolumn{2}{c|}{\textbf{General VQA}} & \multicolumn{3}{c|}{\textbf{Multimodal Reasoning}} & \multicolumn{1}{c|}{\textbf{Text Rec.}} & \multirow{2}{*}{\textbf{AVG.}} \\
    \cmidrule(lr){2-3} \cmidrule(lr){4-6} \cmidrule(lr){7-7}
    \textbf{Short Data}& \multicolumn{1}{c}{\footnotesize MMBench} & \multicolumn{1}{c|}{\footnotesize RWQA} & \multicolumn{1}{c}{\footnotesize MMMU} & \multicolumn{1}{c}{\footnotesize MMMU-Pro} & \multicolumn{1}{c|}{\footnotesize MathVista} & \multicolumn{1}{c|}{\footnotesize OCRBench} & \\
    \midrule
    Qwen2.5-VL-7B & 80.68 & 68.76 & 53.00 & \textbf{37.80} & \textbf{70.50} & \textbf{88.10} & 66.47 \\
    \midrule
    0\% & 80.00 & \textbf{72.68} & 49.33 & 36.65 & 68.70 & 85.50 & 65.48 \\
    20\% & \textbf{81.82} & 70.98 & \textbf{54.33} & 36.47 & 68.30 & 87.30 & \textbf{66.53} \\
    40\% & 81.25 & 70.46 & 52.67 & 37.05 & 68.30 & 87.10 & 66.14 \\
    60\% & 81.66 & 69.93 & 52.67 & 37.51 & 67.80 & 86.70 & 66.05 \\
    80\% & 81.14 & 69.67 & 52.11 & 37.40 & 69.30 & 87.40 & 66.17 \\
    \bottomrule
    \end{tabular}
    }
\end{table*}

\subsection{Multi-Task Long-Context Data Mixture}
\label{sec:mixing:ratio}
Data composition is critical for long-context extension~\citep{fu2024data,gao2025prolong}.
So far, our studies have evaluated each training task of long-document VQA data in isolation.
We therefore study how to combine these three types of data into a single training mixture.
Specifically, we group them into two categories: information extraction, which combines \extractsingle and \extractmulti evenly, and reasoning, which corresponds to \reasoningtask.

We grid search the extraction-to-reasoning ratio in 20\% increments, ranging from all reasoning (0:10) to all extraction (10:0).
As shown in \cref{tab:extract_reason_ratio}, moderately extraction-heavy mixtures perform best, with 6:4 and 8:2 achieving the highest overall scores.
The best performance is obtained with an extraction-to-reasoning ratio of 8:2, which also outperforms the best single-data setting in \cref{tab:vqa_effectiveness}.
This suggests that combining complementary training tasks is more effective than relying on any single task alone.
It also indicates that retrieving key information from long-context inputs remains a major bottleneck when extending the context window, while retaining a small amount of reasoning data helps preserve task diversity.
We therefore use the 8:2 mixture in subsequent experiments.

\subsection{Short-Context Performance Preservation}
\label{sec:mixing:short}
\begin{wrapfigure}{r}{0.46\textwidth}
    \vspace{-20pt}
    \centering
    \includegraphics[width=0.45\textwidth]{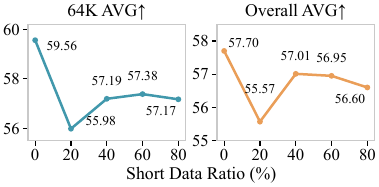}
    \vspace{-5pt}
    \caption{Long-document VQA performance under different short-data mixing ratios. We report 64K AVG and the overall AVG across 64K and 128K. Full results, including 128K scores, are provided in \cref{app:complementary_experiment_results:short_mix_long_vqa}.}
    \label{fig:short_mix_long_vqa_avg}
    \vspace{-8pt}
\end{wrapfigure}
The degradation of short-context capabilities is a common concern in long-context continued pre-training.
To examine this trade-off, we mix different proportions of short-context data into the LongPT stage while keeping the total token budget fixed.
We obtain the short-context data from LLaVA-OneVision~\citep{li2024llava}, with details provided in \cref{app:short_context_data_details}.
We keep the 8:2 extraction-to-reasoning mixture fixed as the long-context component and vary the short-context data ratio from 0\% to 80\% in increments of 20\%.
We summarize the averages of long-context performance in \cref{fig:short_mix_long_vqa_avg}, with full results provided in \cref{app:complementary_experiment_results:short_mix_long_vqa}.
In particular, we report short-context performance on six benchmarks across three capabilities in \cref{tab:short_mix_short_performance}, with evaluation details provided in \cref{app:evaluation_details:short_context}.

\noindent\textbf{Results and trade-off.}
A somewhat surprising observation is that pure long-context training does not severely degrade short-context capabilities.
With 0\% short data, the model achieves the best long-document VQA average of 57.70, with only a mild drop in short-context average from 66.47 to 65.48.
This suggests that high-quality long-document VQA data can preserve the model's general short-context ability, possibly because its QA format still follows an instruction-following style despite the substantially longer input context.

Meanwhile, short-context mixing introduces a clear trade-off.
Adding 20\% short-context data yields the best short-context average of 66.53, but lowers the long-document VQA score to 55.57.
In contrast, the 40\% setting provides a better practical balance: it better preserves long-document performance, averaging 57.01, while keeping short-context performance close to the original model at 66.14.

Overall, since our goal is to maximize long-context capability without substantial degradation in short-context performance, we adopt pure long-context training without short-context data in the final recipe.
The 40\% setting can be viewed as a balanced alternative when stronger short-context preservation is required.

%\vspace{-5pt}
\section{\modelname Performance and Generalization}
\label{sec:final_results}

\begin{table*}[t]
    \centering
    \caption{\textbf{Final long-document VQA results.} We compare \modelname, trained with the final LongPT recipe, against representative open-source and closed-source LVLMs at 64K and 128K.
    }
    \label{tab:final_longdoc_results}
    \setlength{\tabcolsep}{4pt}
    \resizebox{0.9\textwidth}{!}{%
    \begin{tabular}{l*{4}{C{1.2cm}}|*{4}{C{1.2cm}}|C{1.2cm}}
    \toprule
    \multirow{2}{*}{\textbf{Model}} & \multicolumn{4}{c|}{\textbf{64K MMLongBench}} & \multicolumn{4}{c|}{\textbf{128K MMLongBench}} & \multirow{2}{*}{\textbf{AVG.}} \\
    \cmidrule(lr){2-5} \cmidrule(lr){6-9}
    & \multicolumn{1}{c}{\footnotesize MMLB-D} & \multicolumn{1}{c}{\footnotesize LD-URL} & \multicolumn{1}{c}{\footnotesize SLIDE} & \multicolumn{1}{c|}{\footnotesize AVG.} & \multicolumn{1}{c}{\footnotesize MMLB-D} & \multicolumn{1}{c}{\footnotesize LD-URL} & \multicolumn{1}{c}{\footnotesize SLIDE} & \multicolumn{1}{c|}{\footnotesize AVG.} & \\
    \midrule
    \multicolumn{10}{l}{\textbf{Small open-source LVLMs ($<15$B)}} \\
    \addlinespace[1pt]
    \modelname (ours) & \textbf{36.00} & \textbf{62.69} & \textbf{80.00} & \textbf{59.56} & \textbf{34.19} & \textbf{56.33} & \textbf{77.00} & \textbf{55.84} & \textbf{57.70} \\
    Qwen2.5-VL-7B & 32.17 & 49.57 & 75.00 & 52.24 & 26.96 & 51.85 & 68.00 & 48.94 & 50.59 \\
    InternVL3-8B & 27.72 & 52.84 & 70.00 & 50.19 & 25.89 & 48.44 & 58.00 & 44.11 & 47.15 \\
    InternVL3-14B & 35.33 & 49.69 & 67.00 & 50.67 & 29.61 & 50.21 & 53.00 & 44.27 & 47.47 \\
    InternVL3.5-8B & 29.26 & 47.94 & 38.00 & 38.40 & 18.28 & 31.89 & 0.00 & 16.72 & 27.56 \\
    InternVL3.5-14B & 29.04 & 48.57 & 53.00 & 43.54 & 16.91 & 33.86 & 0.00 & 16.92 & 30.23 \\
    Gemma3-4B & 25.00 & 33.33 & 57.00 & 38.44 & 26.44 & 33.38 & 53.00 & 37.61 & 38.03 \\
    Gemma3-12B & 31.52 & 49.59 & 63.00 & 48.03 & 30.56 & 51.91 & 60.00 & 47.49 & 47.76 \\
    Gemma4-E2B & 28.00 & 29.86 & 48.00 & 35.29 & 20.83 & 27.73 & 51.00 & 33.19 & 34.24 \\
    Gemma4-E4B & \textbf{36.00} & 30.21 & 45.00 & 37.07 & 28.50 & 37.78 & 45.00 & 37.09 & 37.08 \\
    \midrule
    \multicolumn{10}{l}{\textbf{Large open-source LVLMs ($\geq15$B)}} \\
    \addlinespace[1pt]
    Qwen2.5-VL-32B & 44.06 & 60.58 & 77.00 & 60.55 & 32.08 & 60.15 & 76.00 & 56.08 & 58.31 \\
    Qwen2.5-VL-72B & 51.97 & 58.96 & 77.5 & 62.81 & 40.19 & \textbf{62.45} & 73.9 & 58.85 & 60.83 \\
    InternVL3-38B & 36.00 & 49.32 & 74.00 & 53.11 & 29.65 & 50.33 & 54.00 & 44.66 & 48.88 \\
    InternVL3.5-38B & 39.17 & 48.37 & 52.00 & 46.51 & 22.24 & 40.09 & 1.01 & 21.11 & 33.81 \\
    Gemma3-27B & 38.00 & 51.72 & 69.00 & 52.91 & 27.94 & 61.13 & 68.00 & 52.35 & 52.63 \\
    Gemma4-26B-A4B & 40.67 & 42.82 & 64.00 & 49.16 & 35.68 & 43.25 & 60.00 & 46.31 & 47.74 \\
    Gemma4-31B & \textbf{56.33} & \textbf{62.42} & \textbf{84.00} & \textbf{67.58} & \textbf{51.87} & 62.00 & \textbf{86.00} & \textbf{66.62} & \textbf{67.10} \\
    \midrule
    \multicolumn{10}{l}{\textbf{Closed-source LVLMs}} \\
    \addlinespace[1pt]
    GPT-5.4 & 61.56 & 72.37 & 87.10 & 73.68 & 52.96 & 73.06 & N/A & 63.01 & 69.41 \\
    GPT-5.5 & \textbf{93.10} & \textbf{92.52} & \textbf{95.77} & \textbf{93.80} & \textbf{83.33} & \textbf{89.12} & N/A & \textbf{86.22} & \textbf{90.77} \\
    Gemini-2.5-Pro & 65.40 & 68.13 & 89.00 & 74.18 & 65.04 & 67.60 & 91.00 & 74.55 & 74.37 \\
    Gemini-3.1-Flash & 74.43 & 79.32 & 91.00 & 81.58 & 69.63 & 76.38 & 90.00 & 78.67 & 80.13 \\
    Gemini-3.1-Pro & 79.40 & 78.25 & 93.00 & 83.55 & 77.75 & 80.57 & \textbf{93.00} & 83.77 & 83.66 \\
\bottomrule
\end{tabular}
    }
\vspace{-4pt}
\end{table*}

\begin{wrapfigure}{r}{0.45\textwidth}
    \vspace{-15pt}
    \centering
    \includegraphics[width=0.43\textwidth]{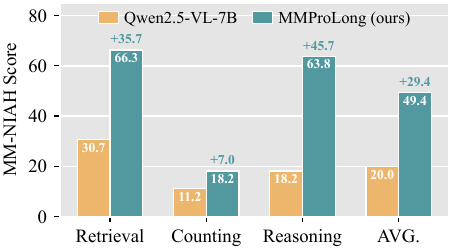}
    \vspace{-4pt}
    \caption{\textbf{MM-NIAH Scores.} We report scores averaged over 64K and 128K contexts for retrieval, counting, reasoning, and the overall average. Additional baselines are provided in \cref{app:complementary_experiment_results:generalization_niah}.}
    \label{fig:generalization_niah}
\vspace{-10pt}
\end{wrapfigure}

Based on the above design choices, we arrive at the final training recipe of \modelname, with the full configuration in \cref{app:final_recipe}.
In this section, we first compare \modelname with a wide range of LVLM baselines on long-document VQA.
Besides this, we further show that \modelname can generalize to
\begin{inparaenum}[(i)]
\item longer contexts up to 512K without further training or adaptation;
\item diverse long-context tasks without task-specific training, including MM-NIAH for webpage-based needle-in-a-haystack evaluation, VTCBench for long-context vision-text compression, and long-video understanding benchmarks.
\end{inparaenum}
To further evaluate the generalizability of our training recipe, we also validate it on Qwen3-VL-8B, with results shown in \cref{app:complementary_experiment_results:backbone}.

\subsection{\modelname Compared with LVLM Baselines}
\label{sec:final_results:longdoc}

We compare \modelname, trained with the final recipe, against a wide range of open-source and closed-source LVLM baselines, as shown in \cref{tab:final_longdoc_results}.
The full list of evaluated models is provided in \cref{app:evaluation_details:final_recipe}.
Among open-source LVLMs below 15B parameters, \modelname achieves the best overall average of 57.70, improving the Qwen2.5-VL-7B base model from 50.59 by 7.11\%.
The improvement is consistent across both context lengths, raising the average score from 52.24 to 59.56 at 64K and from 48.94 to 55.84 at 128K.
Notably, \modelname also outperforms several substantially larger open-source LVLMs, including InternVL3-38B and Gemma3-27B, achieving competitive long-context performance among open-source models.

\subsection{Generalization Beyond the Training Setting}
\label{sec:final_results:generalization}

\noindent\textbf{Generalization to longer contexts.} In this experiment, we examine whether the final \modelname recipe can extrapolate beyond its 128K training context.
To this end, we further extend the long-document VQA benchmarks to 256K and 512K contexts.
Following MMLongBench~\citep{wang2025mmlongbench}, we construct these longer contexts by alternately padding the left and right sides with randomly sampled negative documents, while keeping the original questions and relevant evidence unchanged.
Further details are provided in \cref{app:evaluation_details:longer_context}.
As shown in \cref{tab:generalization_longer_context}, \modelname generalizes to longer contexts without additional training or adaptation.
While Qwen2.5-VL-7B degrades sharply as the context length increases, dropping from 38.12 at 256K to 19.49 at 512K, \modelname maintains strong performance at both lengths, achieving 55.09 at 256K and 52.52 at 512K.
This results in a significant overall advantage over the original Qwen2.5-VL-7B model, increasing the average score from 28.80 to 53.80.

\begin{table*}[t]
\centering
\caption{We evaluate the final 128K \modelname at 256K and 512K without additional training or adaptation.
MMLongBench-Doc (MMLB-D), LongDocURL (LD-URL), and SlideVQA (SLIDE) are reported in order. Some baselines fail on 512K SlideVQA due to the large number of images.
}
\label{tab:generalization_longer_context}
\setlength{\tabcolsep}{4pt}
\resizebox{0.95\textwidth}{!}{%
\scriptsize
\begin{tabular}{l*{4}{C{1.05cm}}|*{4}{C{1.05cm}}|C{1.05cm}}
\toprule
\multirow{2}{*}{\textbf{Model}} & \multicolumn{4}{c|}{\textbf{256K MMLongBench}} & \multicolumn{4}{c|}{\textbf{512K MMLongBench}} & \multirow{2}{*}{\textbf{AVG.}} \\
\cmidrule(lr){2-5} \cmidrule(lr){6-9}
& \multicolumn{1}{c}{\footnotesize MMLB-D} & \multicolumn{1}{c}{\footnotesize LD-URL} & \multicolumn{1}{c}{\footnotesize SLIDE} & \multicolumn{1}{c|}{\footnotesize AVG.} & \multicolumn{1}{c}{\footnotesize MMLB-D} & \multicolumn{1}{c}{\footnotesize LD-URL} & \multicolumn{1}{c}{\footnotesize SLIDE} & \multicolumn{1}{c|}{\footnotesize AVG.} & \\
\midrule
\modelname & 29.69 & \textbf{58.58} & \textbf{77.00} & \textbf{55.09} & \textbf{31.91} & \textbf{55.65} & \textbf{70.00} & \textbf{52.52} & \textbf{53.80} \\
Qwen2.5-VL-7B & 25.47 & 35.88 & 53.00 & 38.12 & 13.44 & 24.61 & 20.41 & 19.49 & 28.80 \\
Gemma3-4B & 20.89 & 32.68 & 44.00 & 32.52 & 20.39 & 26.14 & 0.00 & 15.51 & 24.02 \\
Gemma3-12B & \textbf{31.63} & 47.47 & 63.00 & 47.37 & 24.18 & 46.37 & 0.00 & 23.51 & 35.44 \\
\bottomrule
\end{tabular}
}
\end{table*}

\begin{wrapfigure}{r}{0.45\textwidth}
    \vspace{-13pt}
    \centering
    \includegraphics[width=0.42\textwidth]{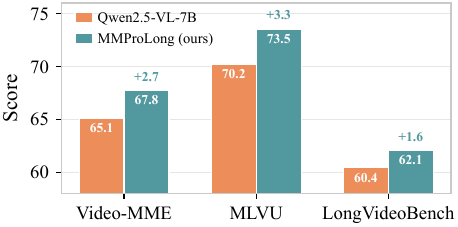}
    \vspace{-4pt}
    \caption{\textbf{Long-video generalization.} We report scores on Video-MME, MLVU, and LongVideoBench. Full results are provided in \cref{app:complementary_experiment_results:generalization_video}.}
    \label{fig:generalization_video}
    \vspace{-17pt}
\end{wrapfigure}

Having shown that the final recipe improves long-document VQA performance and extrapolates to longer contexts, we next examine whether the learned long-context capability transfers to other multimodal tasks.
\Cref{fig:generalization_niah,fig:generalization_video} report results on MM-NIAH and long-video understanding benchmarks, namely Video-MME~\citep{fu2025video}, MLVU~\citep{zhou2025mlvu}, and LongVideoBench~\citep{wu2024longvideobench}.
We also evaluate long-context vision-text compression on VTCBench~\citep{zhao2025vtcbench}, with full results provided in \cref{app:complementary_experiment_results:generalization_vtc}.

MM-NIAH~\citep{wang2024needle} is a multimodal needle-in-a-haystack benchmark that evaluates retrieval, counting, and reasoning tasks over webpage-based haystacks. As shown in \cref{fig:generalization_niah}, \modelname substantially improves over Qwen2.5-VL-7B, increasing the average score from 20.0 to 49.4.
The gains are especially pronounced on retrieval and reasoning tasks, suggesting that long-document VQA training improves the model's ability to locate sparse evidence and use it for reasoning in long multimodal contexts.

We observe similar transfer on long-video understanding benchmarks and VTCBench.
On long-video benchmarks in \cref{fig:generalization_video}, \modelname consistently improves over Qwen2.5-VL-7B on Video-MME, MLVU, and LongVideoBench, despite not using video-specific training data; detailed scores are provided in \cref{app:complementary_experiment_results:generalization_video}.
Meanwhile, VTCBench evaluates LVLMs' ability to perform long-context vision-text compression across three tasks: retrieval, reasoning, and memory.
As shown in \cref{tab:generalization_vtc} in \cref{app:complementary_experiment_results:generalization_vtc}, \modelname improves the overall score on VTCBench from 48.23 to 52.73, with gains on both reasoning and memory tasks while maintaining strong retrieval performance.
Together, these results indicate that our proposed LongPT recipe learns a general long-context multimodal capability rather than overfitting to the document VQA format.

\section{Conclusion}
In this work, we presented a systematic study of long-context continued pre-training for LVLMs, focusing on how to construct and mix effective multimodal long-context data.
Our experiments show that long-document VQA is a strong and practical training task, as it provides diverse retrieval and reasoning supervision signals, preserves short-context capability, and supports context extension from 32K to 128K under a modest token budget.
Instantiated as \modelname, our recipe improves long-document VQA performance and generalizes beyond the training context window to 256K and 512K context lengths, as well as broader multimodal long-context tasks, such as long-video understanding.
We hope our study provides a practical foundation for building future LVLMs with reliable long-context capability.

\section*{Acknowledgments}
The authors of this paper were supported by the National Key Research and Development Program of China (2025YFE0200500), the ITSP Platform Research Project (ITS/189/23FP) from ITC of Hong Kong, SAR, China, and the AoE (AoE/E-601/24-N), the CRF (No. C6004-25G), the RIF (R6021-20) and the GRF (16205322) from RGC of Hong Kong, SAR, China.
\clearpage

\clearpage

\bibliographystyle{unsrtnat}
\bibliography{reference}

\clearpage
\beginappendix
\crefname{appendix}{Appendix}{Appendices}
\Crefname{appendix}{Appendix}{Appendices}
\crefalias{section}{appendix}
\crefalias{subsection}{appendix}
\crefalias{subsubsection}{appendix}
\section{Final Recipe and Implementation}
\label{app:final_recipe_and_implem}
\subsection{Final LongPT Recipe}
\label{app:final_recipe}
We summarize the final LongPT recipe used for the main results.
The recipe is derived from the design choices studied in \cref{sec:curation,sec:mixing}: we use long-document VQA as the primary source of synthesizing data, naturally sample training sequences over the target length range of 32K--128K, and use an $8:2$ extraction-to-reasoning mixture. We list the full configuration of the final LongPT recipe in \cref{tab:final_recipe}.

\begin{table}[h]
    \centering
    \small
    \caption{\textbf{Final training recipe for \modelname}, which is selected from the ablation studies in \cref{sec:curation,sec:mixing}.}
    \label{tab:final_recipe}
    \setlength{\tabcolsep}{5pt}
    \resizebox{0.9\textwidth}{!}{%
    \begin{tabular}{lll}
        \toprule
        \multicolumn{3}{c}{\textbf{Long-Context Continued Pre-Training (LongPT)}} \\
        \midrule
        \textbf{Long Data}
            & Data Synthesis & \extractsingle, \extractmulti, and \reasoningtask \\
            \addlinespace
            & Distribution & \emph{Pool-native}: natural sampling from the document pool \\
            & & over the target length range $[32\mathrm{K},128\mathrm{K}]$ \\
            \addlinespace
& Mixture & 8:2 extraction-to-reasoning ratio. The three task ratio is \\
            & & 40\%\extractsingle, 40\%\extractmulti, 20\%\reasoningtask \\
        \addlinespace
        \textbf{Short Data}
            & Default & None (pure long-context data) \\
            \addlinespace
            & Alternative & $60\%$ long-context data, $40\%$ LLaVA-OneVision data \\
            & & for better short-context preservation \\
        \addlinespace
        \textbf{Model} & Initialization & Qwen2.5-VL-7B-Instruct (original mRoPE base freq. $1\times10^6$) \\
        & mRoPE for 128K & $4\times 10^6$ \\
        & Maximum length & 131,072 (128K tokens) \\
        \addlinespace
        \textbf{Optim.}
            & Token Budget & 5B tokens (2.9K H20 hours) \\
            & Optimizer & AdamW (weight decay = 0.1, $\beta_1$ = 0.9, $\beta_2$ = 0.95) \\
            & LR schedule & $1\times10^{-5}$ with 10\% warmup and cosine decay to $1\times10^{-6}$ \\
            & Batch size & 4M tokens (32 sequences) \\
            & Framework & VeOmni with FlashAttention \\
            & Parallelism & Sequence parallelism size 2 and FSDP size 4 \\
        \bottomrule
    \end{tabular}
    }
\end{table}

\subsection{Training Implementation Details}
\label{app:final_recipe:implementation_details}
We conduct our LongPT experiments based on Qwen2.5-VL-7B~\citep{bai2025qwen2}, whose original context window is 32K, and extend it to 128K. 
For RoPE base frequency, we follow Dynamic-NTK~\citep{dynamicntk} and set the mRoPE base frequency from the original value of $1\times 10^6$ to $4\times 10^6$ by default.
We present the ablation study of the base frequency in \cref{app:complementary_experiment_results:mrope_base}.
Then, we further apply the final training recipe to Qwen3-VL-8B~\citep{bai2025qwen3} to test whether the recipe transfers across backbone models in \cref{app:complementary_experiment_results:backbone}.
We train both models with VeOmni~\citep{ma2025veomni}, a scalable framework for multimodal pre-training.
The optimizer used is AdamW, using a peak learning rate of $1\times10^{-5}$, cosine decay to $1\times10^{-6}$, and $10\%$ linear warmup.
Each LongPT run is trained for a fixed budget of $5B$ tokens, with a maximum sequence length of $131{,}072$ tokens (128K) and a global batch size of 4M tokens, which contains 32 sequences per update. 
Throughout the paper, we use binary prefixes: $K=2^{10}$, $M=2^{20}$, and $B=2^{30}$.

To speed up, we use FlashAttention~\citep{daoflashattention} for efficient attention computation in long-context data. In addition, Ulysses sequence parallelism size $2$ and FSDP size $4$ are used to fit the 128K training configuration on a single 8-GPU NVIDIA H20 node; in practice, we train with 8 H20 nodes (64 GPUs in total) to improve throughput.

\section{Full Evaluation Details}
\label{app:evaluation_details}
We describe the details of our evaluation. Here, for tasks from MMLongBench~\citep{wang2025mmlongbench}, including long-document VQA and MM-NIAH tasks, we follow their evaluation and use their released code v1.1~\footnote{\url{https://github.com/EdinburghNLP/MMLongBench}}.
Meanwhile, we use the VLMEvalKit~\citep{duan2024vlmevalkit} to evaluate LVLMs on other tasks, including VTCBench, long video benchmarks, and short-context benchmarks.

\subsection{Long-Document VQA}
\label{app:evaluation_details:longdoc}
For the main long-document evaluation, we use the document category of MMLongBench~\citep{wang2025mmlongbench}, which contains MMLongBench-Doc~\citep{ma2024mmlongbenchdoc}, LongDocURL~\citep{deng2025longdocurl}, and SlideVQA~\citep{tanaka2023slidevqa}.
All examples in this category are instantiated at five standardized context lengths: 8K, 16K, 32K, 64K, and 128K tokens.
Unless otherwise specified, we evaluate each model at 64K and 128K context lengths.

For grading, we follow the MMLongBench v1.1 evaluation protocol and report the official LLM-judged document QA score for each dataset.
Specifically, this v1.1 protocol introduces LLM-based judging for the long-document VQA tasks, which handles different answer formats separately.
For simple answer formats, such as string, integer, float answers, and ``not answerable,'' the judge assigns a binary score indicating whether the predicted answer matches the reference answer as shown in \cref{tab:binary_judge_prompt}.
For list-style answers, the evaluator first extracts the predicted list and then computes an F1 score based on the overlap between the predicted list and the reference list. We show this prompt in \cref{tab:list_judge_prompt}.
For each context length, the AVG. column is the macro average over the three datasets, and the overall AVG. is the macro average over the 64K and 128K results.

\begin{table}[h]
    \centering
    \caption{Prompt for the binary answer judge in the long-document VQA category of MMLongBench. Gray highlighted spans denote placeholders filled at evaluation time.}
    \label{tab:binary_judge_prompt}
    \begin{tabular}{p{0.96\linewidth}}
    \toprule
    \begin{minipage}{0.96\linewidth}
    \raggedright\ttfamily\footnotesize

    Now your role is a grading teacher. Your task is to review and score student answers based on reference standard answers. You need to notice the following key points:\\[2pt]
    - First, extract the final answer from the student's solution, then analyze and judge whether the answer is correct.\\
    - Scoring should only refer to the final answer obtained by the student; there is no need to examine whether the intermediate problem-solving steps are correct.\\
    - When analyzing and judging whether the answer is correct, you need to write down the scoring rationale, organize it into clear statements that follow the logical flow. The summary of the scoring rationale should be placed at the end, using the following format: "In summary, the student's answer deserves x points" (where x represents the student's specific score). \\
    - Keep the whole process concise, within 150 words.\\
    - Provide the score based on your analysis and display it in a code block in "JSON" format.\\
    - An item is covered if it is strictly mentioned or unambiguously implied by a semantic equivalence. This includes numerical equivalence (e.g., 10\% and 0.1), synonyms (e.g., UK and United Kingdom), and plural/singular forms (e.g., "apple" and "apples"). However, do not accept loosely related concepts.\\[2pt]
    Your output format is:\\
    {[}Scoring Rationale{]}:\\
    {[}Score{]}: x points\\
    {[}JSON{]}:\\
    \{"answer\_score": \textless integer\_value\textgreater\}\\[2pt]
    \quad \\
    Below is the grading rubric:\\
    {[}Scores{]}: \\
    The scoring scale consists of 2 levels in total, from highest to lowest: 1 point, 0 points (the minimum is 0 points; if a situation arises where points need to be deducted beyond 0, simply assign 0 points).\\
    {[}Tier Details{]}: \\
    1 point: Assign 1 point if the student's final answer matches the standard answer. If the question has multiple sub-questions, all sub-questions must be answered to assign 1 point.\\
    0 points: Assign 0 points if the student's final answer does not match the standard answer.
    \quad \\
    \quad \\
    \promptplaceholder{<in-context exemplars\_1>} \\
    \quad \\
    \promptplaceholder{<in-context exemplars\_2>} \\
    \quad \\
    \promptplaceholder{<in-context exemplars\_3>} \\
    \quad \\
    \promptplaceholder{<test\_case\_1>} \\
    \end{minipage} \\
    \bottomrule
    \end{tabular}
\end{table}

\begin{table}[h]
    \centering
    \caption{Prompt for the list-style answer judge in the long-document VQA category of MMLongBench.}
    \label{tab:list_judge_prompt}
    \begin{tabular}{p{0.96\linewidth}}
    \toprule
    \begin{minipage}{0.96\linewidth}
    \raggedright\ttfamily\footnotesize
    Now your role is a grading teacher. Your task is to review and score student answers for LIST-style questions, where the standard answer is a list of required items.\\
    - First, extract the specific list of items from the \textless Student Answer\textgreater. Ignore conversational filler (e.g., "The answer is...").\\
    - Then, compare the {[}Extracted List{]} against the \textless Standard Answer\textgreater\ (Ground Truth).\\[2pt]
    Here are some extra key points:\\
    - The standard answer is a JSON-like list of items with each item as one required element. Determine whether each item is covered by the student's answer list.\\
    - An item is covered if it is strictly mentioned or unambiguously implied by a semantic equivalence. This includes numerical equivalence (e.g., 10\% and 0.1), synonyms (e.g., UK and United Kingdom), and plural/singular forms (e.g., "apple" and "apples"). However, do not accept loosely related concepts.\\
    - You need to write down the extraction and comparing rationale, organize it into clear statements that follow the logical flow. The summary of the rationale should be placed at the end, using the following format: "In summary, the student's answer list has X items, covering Y items from the reference list."\\
    - Keep the whole process concise, within 200 words.\\
    - Provide the student's answer item count and covered item count in a code block in "JSON" format.\\[2pt]
    Your output format is:\\
    {[}Rationale{]}:\\
    {[}JSON{]}:\\
    \{\\
    \hspace*{1em}"student\_answer\_count": \textless integer\_value\textgreater,\\
    \hspace*{1em}"covered\_count": \textless integer\_value\textgreater\\
    \}
    \quad \\
    \quad \\
    \promptplaceholder{<in-context exemplars\_1>} \\
    \quad \\
    \promptplaceholder{<in-context exemplars\_2>} \\
    \quad \\
    \promptplaceholder{<in-context exemplars\_3>} \\
    \quad \\
    \promptplaceholder{<test\_case\_1>} \\
    \end{minipage} \\
    \bottomrule
    \end{tabular}
\end{table}

\subsection{Question-Type Composition}
\label{app:evaluation_details:question_types}
We summarize the composition of the long-document VQA evaluation from two perspectives in \cref{tab:question_type_composition}.
LongDocURL provides official \texttt{task\_tag} labels for understanding, reasoning, and locating questions.
We also partition each dataset by annotated evidence pages and answer format: after separating unanswerable questions, we group answerable questions with multiple evidence pages as cross-page, and divide the remaining single-page questions into text, numeric, and list answers.
The percentages below are reported for the 128K evaluation set.
The 256K and 512K variants reuse the same questions and relevant evidence, so their question-type composition is unchanged by the negative-document padding.

\begin{table}[t]
    \centering
    \small
    \caption{\textbf{Question-type composition of the long-document VQA evaluation.} (a) Official LongDocURL task labels. (b) Mutually exclusive groups based on evidence scope and answer format. Values are percentages of questions at 128K; each dataset sums to 100\%. MMLB-D, LD-URL, and SLIDE denote MMLongBench-Doc, LongDocURL, and SlideVQA, respectively.}
    \label{tab:question_type_composition}
    \setlength{\tabcolsep}{6pt}
    \textbf{(a) LongDocURL official task labels}\\[3pt]
    \begin{tabular}{ccc}
        \toprule
        Understanding & Reasoning & Locating \\
        \midrule
        63\% & 30\% & 7\% \\
        \bottomrule
    \end{tabular}

    \medskip
    \textbf{(b) Evidence scope and answer format}\\[3pt]
    \begin{tabular}{lrrr}
        \toprule
        Question group & MMLB-D & LD-URL & SLIDE \\
        \midrule
        Single-page text & 17\% & 15\% & 50\% \\
        Single-page numeric & 27\% & 25\% & 37\% \\
        Single-page list & 6\% & 8\% & 0\% \\
        Cross-page evidence & 28\% & 51\% & 13\% \\
        Unanswerable & 22\% & 1\% & 0\% \\
        \bottomrule
    \end{tabular}
\end{table}

\subsection{Details of Long-Document VQA Evaluation with the Final Recipe}
\label{app:evaluation_details:final_recipe}
In evaluating \modelname trained with the final recipe, we compare it with a wide range of open- and closed-source LVLMs.
For open-source models, we evaluate Qwen2.5-VL (7B, 32B, 72B)~\citep{bai2025qwen2}, InternVL3 (8B, 14B, 38B)~\citep{zhu2025internvl3}, InternVL3.5 (8B, 14B, 38B)~\citep{wang2025internvl35}, Gemma3 (4B, 12B, 27B)~\citep{team2025gemma3}, and Gemma4 (E2B, E4B, 26B-A4B, 31B)~\citep{google2026gemma4}.
For closed-source models, we include GPT-5.4~\citep{openai2026introducing_gpt5_4}, GPT-5.5~\citep{openai2026introducing_gpt5_5}, Gemini-2.5-Pro~\citep{google2025gemini2_5}, Gemini-3.1-Flash~\citep{gemini_3_1_pro_2026}, and Gemini-3.1-Pro~\citep{gemini_3_1_pro_2026} with a high reasoning budget.

\subsection{Longer-Context Evaluation up to 512K}
\label{app:evaluation_details:longer_context}
To test whether the 128K recipe extrapolates to longer contexts, we further evaluate the same three long-document VQA datasets at 256K and 512K.
The examples at 256K and 512K context lengths are obtained following MMLongBench~\citep{wang2025mmlongbench}: we alternately pad the left and right sides with randomly sampled negative documents until the required length is reached.
No additional training or inference-time adaptation is applied to tested models when evaluating at these longer lengths.
We use the same official LLM-judged score and macro-average computation as in the main long-document evaluation.

\subsection{MM-NIAH: Evaluation on Webpage Haystacks}
\label{app:evaluation_details:mm_niah}
MM-NIAH~\citep{wang2024needle} is a multimodal needle-in-a-haystack benchmark that builds on the webpages from OBELICS~\citep{laurenccon2023obelics}.
In particular, the benchmark contains three task families: retrieval, counting, and reasoning, each with two variants: text-needle and image-needle.
In our evaluation, we use the standardized version provided by MMLongBench~\citep{wang2025mmlongbench}, where each example is instantiated at five context lengths: 8K, 16K, 32K, 64K, and 128K tokens.
In our study, we report results at 64K and 128K context lengths as we extend the context window from 32K to 128K.
For reporting, we first average the text and image variants within each task family.
Then, the average (AVG.) score is the macro average over the three task-family scores.
We use the metrics provided by MMLongBench~\citep{wang2025mmlongbench} for each task variant, including exact match, soft accuracy, and multiple-choice accuracy, depending on the specific subtask definition.

\subsection{VTCBench: Evaluation on Long-Context Vision-Text Compression}
\label{app:evaluation_details:vtcbench}
VTCBench~\citep{zhao2025vtcbench} is another multimodal long-context benchmark that evaluates whether a model can preserve visual-text information under compressed visual context.
We follow the VTCBench-Wild setting reported in the original paper and report all three task scores: Retrieval, Reasoning, and Memory.
We follow the same grading rules as the original paper.
Retrieval and Reasoning are measured by the official \texttt{containsAll} accuracy, which checks whether all ground-truth answers are contained in the model prediction.
Memory is measured by the official LLM accuracy, where \texttt{gpt-4o-mini} judges answer correctness.
The AVG. score follows the Overall column of VTCBench-Wild and is the sample-count weighted average over the three splits, with 800 retrieval examples, 800 reasoning examples, and 600 memory examples.

\subsection{Long-Video Understanding Evaluation}
\label{app:evaluation_details:video}
For long-video understanding, we report the performance on Video-MME~\citep{fu2025video}, MLVU~\citep{zhou2025mlvu}, and LongVideoBench~\citep{wu2024longvideobench}.
We evaluate these benchmarks using the 1fps variants and cap the maximum number of sampled frames per video at 768, with the total number of video tokens not exceeding 24,576.
This configuration is fully aligned with Qwen2.5-VL technical report~\citep{bai2025qwen2}.
For Video-MME and LongVideoBench, we report overall multiple-choice accuracy.
For MLVU, we evaluate the multiple-choice subset and report its average accuracy.

\subsection{Short-Context Evaluation}
\label{app:evaluation_details:short_context}
To monitor whether \modelname suffers degradation in general VLM ability, we evaluate short-context performance across three capabilities and six benchmarks.
These include general VQA (MMBench-V1.1~\citep{liu2024mmbench} and RealWorldQA~\citep{grok15}), multimodal reasoning (MMMU~\citep{yue2024mmmu}, MMMU-Pro~\citep{yue2025mmmupro}, and MathVista~\citep{lu2023mathvista}), and text recognition (OCRBench~\citep{liu2024ocrbench}).

For MMMU, we use \texttt{MMMU\_DEV\_VAL} in VLMEvalKit, which combines the official MMMU development and validation splits.
For MMMU-Pro, we use \texttt{MMMU\_Pro\_10c}, the standard 10-choice variant.
For MathVista, we use the testmini subset.
We report the official score of each benchmark and compute the short-context AVG. as the macro average over the six benchmark scores.

\section{Preliminary Details}
\subsection{Document Pool Statistics}
\label{app:document_pool_stats}
\begin{figure}[h]
    \centering
    \includegraphics[width=0.7\linewidth]{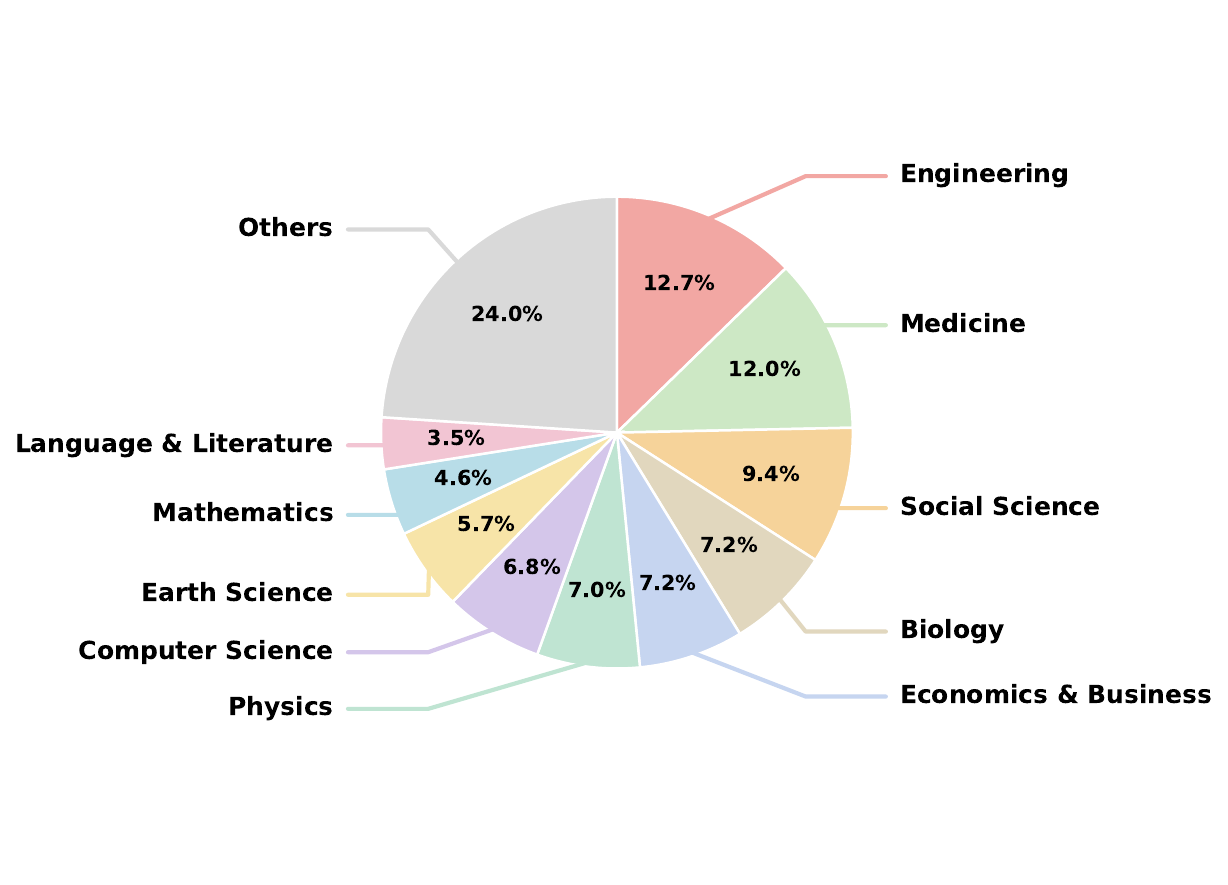}
    \caption{
        The document domain distribution of the document pool used for data synthesis.
    }
    \label{fig:doc_pool_domain_distribution}
\end{figure}

We show the corpus scale, average page length, and language distribution of our document pool in \cref{tab:doc_pool_statistics}. 
Meanwhile, the distribution of the document domain is shown in \cref{fig:doc_pool_domain_distribution}.

\subsection{OCR Expert}
\label{app:ocr_expert}
We use an OCR expert model fine-tuned from Seed~2.0~\citep{seed2_0} to preprocess the rendered PDF pages.
For each page image, the OCR expert parses layout-aware text blocks and assigns structural labels such as title, section heading, paragraph, table, figure caption, header, and footer.

These parsed blocks serve two purposes in our pipeline:
\begin{inparaenum}[(i)]
\item the title and section labels provide lightweight document-structure signals for sampling semantically coherent page spans when generating long-document VQA data;
\item the recognized text blocks provide the target text for constructing OCR transcription baselines.
\end{inparaenum}

\paragraph{OCR capability.}
We evaluate the document-parsing capability of our OCR expert on OmniDocBench~\citep{ouyang2025omnidocbench}.
\Cref{tab:ocr_expert_omnidocbench} reports Overall Edit for English and Chinese documents, where lower scores are better; our OCR expert achieves 0.129 and 0.209, respectively.

\begin{table}[ht]
    \centering
    \small
    \caption{\textbf{OCR model evaluation on OmniDocBench.} Lower Overall Edit scores are better. Our OCR expert supplies the transcription targets used in data synthesis.}
    \label{tab:ocr_expert_omnidocbench}
    \setlength{\tabcolsep}{6pt}
    \begin{tabular}{lcc}
        \toprule
        Model & Overall Edit (EN) $\downarrow$ & Overall Edit (ZH) $\downarrow$ \\
        \midrule
        \multicolumn{3}{l}{\textit{Pipeline tools}} \\
        PaddleOCR-StructureV3 & 0.145 & \textbf{0.206} \\
        MinerU2.1.1-pipeline & 0.162 & 0.244 \\
        \midrule
        \multicolumn{3}{l}{\textit{Expert VLMs}} \\
        MinerU2.0-2505-0.9B & 0.133 & 0.238 \\
        MonkeyOCR-pro-1.2B & 0.146 & 0.221 \\
        OLMOCR-sglang & 0.326 & 0.469 \\
        Our OCR expert & \textbf{0.129} & 0.209 \\
        \bottomrule
    \end{tabular}
\end{table}

\begin{table}[h]
    \centering
    \caption{Statistics of the document pool used for data synthesis.}
    \label{tab:doc_pool_statistics}
    \small
    \setlength{\tabcolsep}{6pt}
    \begin{tabular}{lll}
    \toprule
    \textbf{Category} & \textbf{Statistic} & \textbf{Value} \\
    \midrule
    \multirow{4}{*}{\textbf{Corpus Scale}} 
        & Number of documents & 1,537,504 \\
        & Page-count range & $[20, 200]$ \\
        & Average Pages & 23.80 \\
        & Total pages & 36,592,809 \\
    \midrule
    \multirow{3}{*}{\textbf{Lang.}}
        & English documents & 1,479,370 (96.22\%) \\
        & Chinese documents & 55,202 (3.59\%) \\
        & Other languages & 2,932 (0.19\%) \\
    \bottomrule
    \end{tabular}
\end{table}

\section{Long-Document VQA Training Data Details}
\label{app:long_document_vqa_details}

\subsection{Data Statistics for \emph{Pool-Native} Distribution (Default)}
\label{app:long_document_vqa_details:data_stats}
\begin{figure}[t]
    \centering
    \begin{subfigure}{0.95\linewidth}
        \centering
        \includegraphics[width=\linewidth]{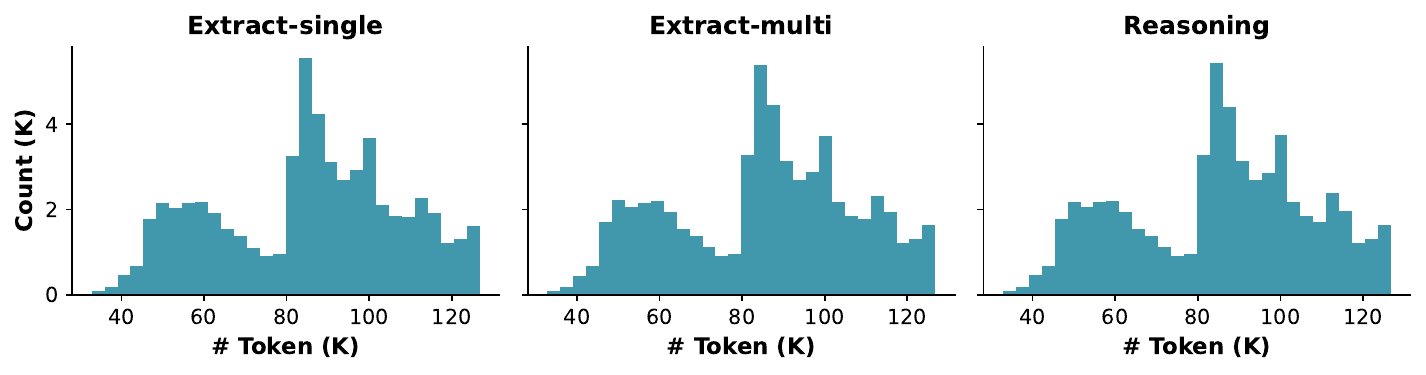}
        \caption{Long-document VQA data in the \emph{pool-native} Distribution (default).}
        \label{fig:longvqa_uniform_token_distribution}
    \end{subfigure}
    \vspace{0.6em}

    \begin{subfigure}{0.95\linewidth}
        \centering
        \includegraphics[width=\linewidth]{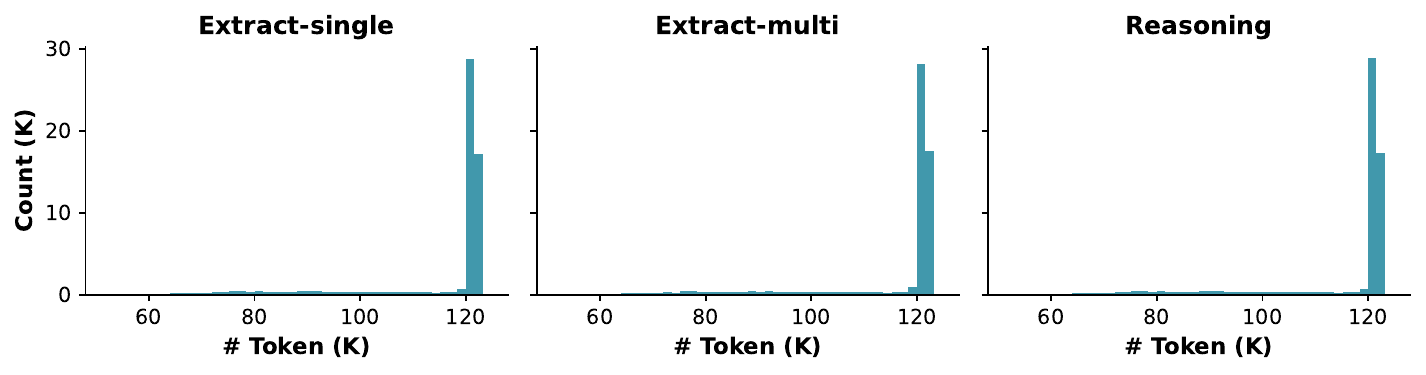}
        \caption{Long-document VQA data in the \emph{long-biased} distribution.}
        \label{fig:longvqa_non_even_token_distribution}
    \end{subfigure}
    \vspace{0.6em}

    \begin{subfigure}{0.95\linewidth}
        \centering
        \includegraphics[width=\linewidth]{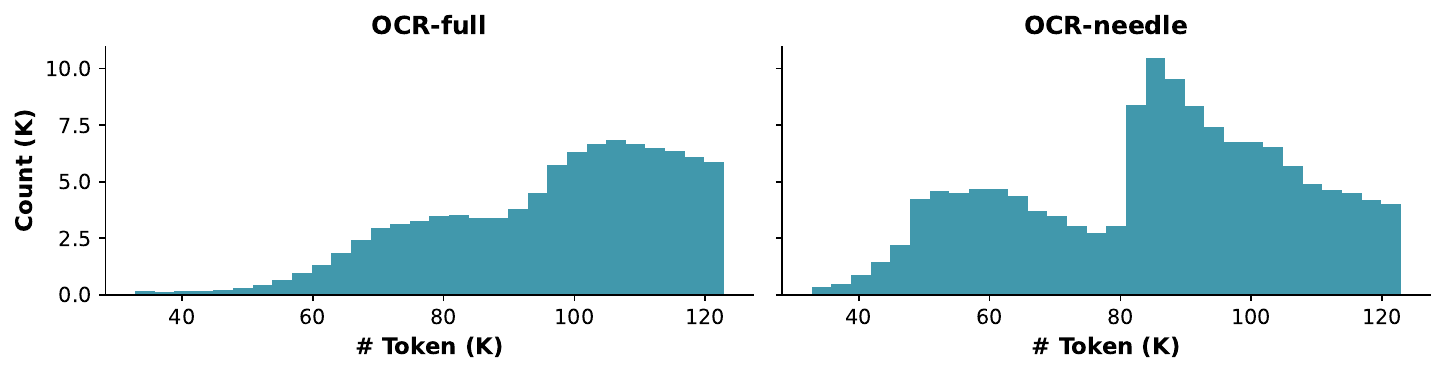}
        \caption{OCR transcription data.}
        \label{fig:ocr_token_distribution}
    \end{subfigure}
    \caption{Token-length distributions of the long-context training data used in our data-design study.}
    \label{fig:long_context_training_data_token_distributions}
\end{figure}

We summarize the statistics of the long-document VQA training data in \cref{tab:longvqa_uniform_data_stats}, with the corresponding token-length distribution shown in \cref{fig:longvqa_uniform_token_distribution}.
This data corresponds to the \emph{pool-native} length distribution used by our final LongPT recipe, covering samples from 32 to 50 rendered PDF pages and approximately 32K--128K multimodal tokens.

\begin{table}[h]
    \centering
    \setlength{\tabcolsep}{2.5pt}
    \caption{Statistics of the long-context training data used in our data-design study.}
    \label{tab:long_context_training_data_stats}
    \small
    \begin{subtable}{\linewidth}
        \centering
        \setlength{\tabcolsep}{2pt}
        \caption{Long-document VQA data in the \emph{pool-native} Distribution (default).}
        \label{tab:longvqa_uniform_data_stats}
        \begin{tabular}{lcccccc}
        \toprule
        \textbf{Task} & \textbf{\# Samples} & \textbf{\# Pages} & \textbf{\# Tokens (K)} & \textbf{Total Tokens (B)} & \textbf{Page Range} & \textbf{Token Range (K)} \\
        \midrule
        \extractsingle & 59{,}055 & 38.4 & 85.3 & 5.04 & $[32, 50]$ & $[32.8, 126.8]$ \\
        \extractmulti & 59{,}316 & 38.4 & 85.3 & 5.06 & $[32, 50]$ & $[32.8, 126.8]$ \\
        \reasoningtask & 59{,}403 & 38.4 & 85.3 & 5.06 & $[32, 50]$ & $[32.8, 126.7]$ \\
        \midrule
        \textbf{Total} & \textbf{177{,}774} & 38.4 & 85.3 & \textbf{15.16} & $[32, 50]$ & $[32.8, 126.8]$ \\
        \bottomrule
        \end{tabular}
    \end{subtable}

    \begin{subtable}{\linewidth}
        \centering
        \setlength{\tabcolsep}{2pt}
        \caption{Long-document VQA data in the \emph{long-biased} distribution.}
        \label{tab:longvqa_non_even_data_stats}
        \begin{tabular}{lcccccc}
        \toprule
        \textbf{Task} & \textbf{\# Samples} & \textbf{\# Pages} & \textbf{\# Tokens (K)} & \textbf{Total Tokens (B)} & \textbf{Page Range} & \textbf{Token Range (K)} \\
        \midrule
        \extractsingle & 59{,}736 & 66.8 & 114.3 & 6.83 & $[50, 100]$ & $[32, 125.3]$ \\
        \extractmulti & 59{,}624 & 66.8 & 114.3 & 6.81 & $[50, 100]$ & $[32, 124.0]$ \\
        \reasoningtask & 59{,}770 & 66.8 & 114.3 & 6.83 & $[50, 100]$ & $[32, 126.7]$ \\
        \midrule
        \textbf{Total} & \textbf{179{,}130} & 66.8 & 114.3 & \textbf{20.47} & $[50, 100]$ & $[32, 126.7]$ \\
        \bottomrule
        \end{tabular}
    \end{subtable}

    \begin{subtable}{\linewidth}
        \centering
        \setlength{\tabcolsep}{2.5pt}
        \caption{OCR transcription data.}
        \label{tab:ocr_data_stats}
        \begin{tabular}{lcccccc}
        \toprule
        \textbf{Task} & \textbf{\# Samples} & \textbf{\# Pages} & \textbf{\# Tokens (K)} & \textbf{Total Tokens (B)} & \textbf{Page Range} & \textbf{Token Range (K)} \\
        \midrule
        \ocrfull & 97{,}336 & 39.6 & 96.4 & 9.38 & $[32, 50]$ & $[32.8, 122.9]$ \\
        \ocrneedle & 140{,}655 & 42.9 & 85.4 & 12.01 & $[32, 50]$ & $[32.8, 122.9]$ \\
        \midrule
        \textbf{Total} & \textbf{237{,}991} & 41.6 & 89.9 & \textbf{21.39} & $[32, 50]$ & $[32.8, 122.9]$ \\
        \bottomrule
        \end{tabular}
    \end{subtable}
\end{table}

\begin{table}[t]
    \centering
    \caption{Percentage of training samples with token length at least 100K for both \emph{pool-native} and \emph{long-biased} distribution.}
    \label{tab:over_100k_ratio}
    \small
    \setlength{\tabcolsep}{8pt}
    \begin{tabular}{lccc}
    \toprule
    \textbf{Distribution} & \textbf{\extractsingle} & \textbf{\extractmulti} & \textbf{\reasoningtask} \\
    \midrule
    pool-native & 23.6\% & 23.6\% & 23.5\% \\
    long-biased & 83.9\% & 83.9\% & 83.9\% \\
    \bottomrule
    \end{tabular}
\end{table}

\subsection{Data Statistics for \emph{Long-Biased} Distribution}
\label{app:long_document_vqa_details:data_stats_long_only}
We summarize the statistics of the \emph{long-biased} long-document VQA training data in \cref{tab:longvqa_non_even_data_stats}.
The corresponding token-length distribution is shown in \cref{fig:longvqa_non_even_token_distribution}.

Compared with the default \emph{pool-native} distribution, this data is sampled from longer documents with 50--100 rendered PDF pages and concentrates more training mass near the upper end of the target context window.
As shown in \cref{tab:over_100k_ratio}, 83.9\% of the samples in the \emph{long-biased} distribution contain at least 100K tokens across all three VQA tasks.
In contrast, only about 23.5\%--23.6\% of the samples in the \emph{pool-native} distribution exceed this threshold.

\subsection{Prompt Template for QA Pair Generation}
\label{app:long_document_vqa_details:prompt}
For each source document, we first use OCR block labels to identify title and section boundaries, then sample a semantically coherent span of 8--15 consecutive pages.
The sampled page images are sent to the teacher LVLM together with a prompt that asks it to synthesize one document-grounded QA pair.

By default, we include two in-context exemplars sampled from a small manually curated exemplar pool.
We instantiate the same core prompt with different task descriptions and page constraints for three data types: single-page extraction, multi-page extraction, and reasoning.

More concretely, \cref{tab:qa_generation_prompt} shows the prompt template used in our data sourcing process, and \cref{tab:qa_generation_task_fields} summarizes the task description and extra restriction inserted for each task type.

\clearpage
\begin{table}[H]
    \centering
    \caption{The prompt template used for long-document QA pair generation. Gray highlighted spans denote placeholders filled dynamically for each sampled document span.}
    \label{tab:qa_generation_prompt}
    \begin{adjustbox}{max width=\textwidth, max totalheight=0.95\textheight}
    \begin{tabular}{p{0.96\linewidth}}
    \toprule
    \begin{minipage}{0.96\linewidth}
    \raggedright\ttfamily\footnotesize
    {[}System{]}\\
    You are an expert in synthesizing document question-answering dialogue. In this task, I will provide you the images of one or more sections from a document and you need to generate a question-answering pair based on the given sections.
    You need to notice the following key points:\\[2pt]
    \quad \\
    {[}Task Definition \& Requirements{]}\\
    The current document QA task is \promptplaceholder{task description}.\\[2pt]
    \quad \\
    {[}General Restrictions{]}\\
    - Questions must be confidently answerable and document-dependent, relying solely on the provided visual content without external knowledge.\\
    - If the question asks about missing information, the absence of such content must be definitively verifiable from the provided image.\\
    - Scope Constraint: Frame questions specifically within the scope of the provided section/pages, strictly avoiding broad references like "across the document" or "the whole file."\\
    - Page Indexing Rule: When writing the evidence description or the "evidence\_pages" field in JSON, strictly use the index provided in the prompt (e.g., use `10' for "Page 10"), regardless of the page number inside images.\\[2pt]
    \promptplaceholder{extra\_restriction} \\
    \quad \\
    {[}Response Format{]}\\
    Please generate the response in two parts: \\
    \quad \\
    Part 1: Evidence Description \\
    Provide a detailed description first, citing specific text or visual elements (using the four evidence types defined below) to support your reasoning. You may use multiple paragraphs if necessary.\\
    \quad \\
    When describing evidence, explicitly categorize the text and visual elements into one of the following four types: \\
    - Text: pure texts, such as paragraphs. \\
    - Layout: text elements with special layout meaning (generalized text), such as titles, headers, footers, table names, and figure names. \\
    - Figure: including charts and general images. \\
    - Table: structured data in rows and columns. \\
    \quad \\
    Part 2: JSON Output \\
    Provide the final question, answer, and metadata in a strict JSON format.\\[2pt]
    \quad \\
    Your answer must strictly fall into one of the following four categories. You must also indicate the type in the JSON output: \\
    1. String: General text, names, short sentences, or short phrases found in the document. \\
        - Example: "John Doe", "Financial Report", "The project was completed on time." \\
   2. Integer: Whole numbers representing counts, years, page numbers, etc. \\
   - Example: "42", "2023", "100". \\
    3. Float: Numbers with decimal points, including currency, percentages, or scientific measurements. \\
   - Example: "12.5", "\$45.20", "98.5\%". \\
    4. List: A collection of multiple items, names, or values. \\
   - Example: ["Apple", "Banana", "Orange"], ["Item A", "Item B"]. \\
   \quad \\
   Your output format is:\\
    {[}Evidence Description{]}:\\
    {[}JSON{]}:\\
    \{"question": \textless question\textgreater, "answer": \textless answer\textgreater, "answer\_format": \textless answer\_format\textgreater, "evidence\_pages": [\textless page\_index\textgreater], "evidence\_sources": [\textless evidence\_sources\textgreater]\}\\[2pt]
    \textless \textgreater\\[2pt]
    \quad \\
    \promptplaceholder{<in-context exemplars\_1>} \\
    \quad \\
    \promptplaceholder{<in-context exemplars\_2>} \\
    \quad \\
    {[}Current Case{]}\\
    The following images represent pages \promptplaceholder{start\_page} to \promptplaceholder{end\_page} of the document. Generate the question-answering pair strictly based on the visual content below, ensuring the question scope is limited to phrases like "the Introduction section", "Pages 20, 21, and 25", or "Page 10." \\
    \promptplaceholder{current\_case\_image\_sequence}
    \end{minipage} \\
    \bottomrule
    \end{tabular}
    \end{adjustbox}
\end{table}
\clearpage

\begin{table}[H]
    \centering
    \caption{The task descriptions and extra restriction for the three long-document VQA tasks. We insert these parts into the prompt template to synthesize data.}
    \label{tab:qa_generation_task_fields}
    \small
    \setlength{\tabcolsep}{4pt}
    \begin{tabular}{L{2.3cm}L{5.6cm}L{6.4cm}}
    \toprule
    \textbf{Task} & \textbf{Task Description} & \textbf{Extra Restriction} \\
    \midrule
    \extractsingle
    & information extraction. You need to generate questions that focus on extracting specific, explicit information directly from the provided document images. Your objective is to create questions that ask for precise facts, entities (such as names, dates, locations, or authors), numerical values, lists of items, or specific steps in a procedure. The answers must be visually present and directly retrievable from the text, tables, or layout elements.
    & Single-Page Focus: Although multiple pages are provided for context, you must generate a question that is **strictly self-contained within a single page**. Select one specific page from the provided sequence and generate a question based solely on its content. The answer must be fully derivable from that single page without requiring cross-referencing or synthesizing information from other pages. \\
    \addlinespace
    \midrule
    \extractmulti
    & information extraction. You need to generate questions that focus on extracting specific, explicit information directly from the provided document images. Your objective is to create questions that ask for precise facts, entities (such as names, dates, locations, or authors), numerical values, lists of items, or specific steps in a procedure. The answers must be visually present and directly retrievable from the text, tables, or layout elements.
    & Multi-Page Priority: When multiple pages are provided, **prioritize** generating questions that require synthesizing information across different pages (>= 2 pages). For example, link a table header from one page to a row on the next, or aggregate data points scattered across multiple pages. You can also come up with other formats of multi-page questions. **Only fall back to single-page questions if no meaningful cross-page connections exist.** \\
    \addlinespace
    \midrule
    \reasoningtask
    & quantitative reasoning. You need to generate questions that require performing calculation, comparison, or counting. Your objective is to create questions involving one of the following: 1. Calculate: Performing arithmetic operations on data found in the document. 2. Compare: Comparing values to identify trends, maximums, minimums, or relative sizes. 3. Count: Counting the frequency of specific items, keywords, or visual elements that satisfy a condition. The answer must require the aforementioned processing step rather than being directly visible as a single contiguous string.
    & - Non-Trivial Synthesis: The question must require **aggregating multiple data points**. For example, ask ``What is the difference in revenue between Q1 and Q2?'' or ``Calculate the total sum of items listed in Table 3.'' The key is that the user must find at least two pieces of information and combine them to get the answer. \\
    \bottomrule
    \end{tabular}
\end{table}

\subsection{Human Verification of QA Pairs}
\label{app:long_document_vqa_details:verification}
We randomly sample 100 generated QA pairs from the three task types and manually verify their answer correctness and evidence consistency.
Among these inspected examples, 97 are fully correct: two contain incorrect answers, and one has an inaccurate evidence annotation.
This suggests that the synthesis pipeline produces high-quality document-grounded supervision overall, while also indicating that the generated data may still contain a small amount of noise.

\section{OCR Transcription Details}
\label{app:ocr_transcription_details}
In addition to VQA-based synthesis, we construct OCR transcription data as a contrasting family of long-context supervision.
Each training example places rendered document pages as visual input and asks the model to generate the corresponding OCR text parsed from those pages.
We consider two variants.
Full-document OCR requires transcribing text elements from all pages of the document, creating dense image-text alignment across the full context.
Needle-page OCR keeps the long visual context but only asks the model to transcribe text from a small subset of selected pages, making the task closer to retrieval over many distractor pages.

\subsection{Data Statistics}
\label{app:ocr_transcription_details:data_stats}
We summarize the statistics of the OCR transcription data in \cref{tab:ocr_data_stats}.
The corresponding token-length distribution is shown in \cref{fig:ocr_token_distribution}.

\section{Short-Context Data Details}
\label{app:short_context_data_details}
We use short-context data in two sets of experiments.
First, in the comparison between long-document VQA and OCR transcription data (\cref{sec:curation:data_method_compare}), we additionally apply supervised fine-tuning (SFT) with short-context instruction data after training the model on OCR transcription data.
Second, in the short-data mixture experiment (\cref{sec:mixing:short}), we mix short-context data with long-context data to study how much short-context supervision should be included in LongPT.

For both settings, we use publicly released short-context instruction data from LLaVA-OneVision~\citep{li2024llava}.
We use the full short-context instruction data in the short-data mixture experiment, while only the SFT portion is used in the additional SFT stage after OCR transcription training.
We download the single-image and multi-image subsets from the LLaVA-OneVision Hugging Face repositories\footnote{\url{https://huggingface.co/datasets/lmms-lab/LLaVA-OneVision-Data}}\footnote{\url{https://huggingface.co/datasets/lmms-lab/M4-Instruct-Data}}.

\section{Complementary Experimental Results}
\label{app:complementary_experiment_results}

\subsection{Sequence-Length Distribution of the Training Data}
\label{app:complementary_experiment_results:length_distribution}
We provide the full dataset-level results for the sequence-length distribution ablation in \cref{tab:length_distribution}.

\begin{table}[H]
    \centering
    \caption{We compare the \emph{pool-native} and \emph{long-biased} distributions under each of the three VQA-based training tasks. We abbreviate these three datasets as MMLB-D, LD-URL, and SLIDE, respectively.}
    \label{tab:length_distribution}
    \setlength{\tabcolsep}{4pt}
    \resizebox{0.95\textwidth}{!}{%
    \begin{tabular}{ll*{4}{C{1.2cm}}|*{4}{C{1.2cm}}|C{1.8cm}}
    \toprule
    & & \multicolumn{4}{c|}{\textbf{64K MMLongBench}} & \multicolumn{4}{c|}{\textbf{128K MMLongBench}} & \multirow{2}{*}{\textbf{AVG.}} \\
     \cmidrule(lr){3-6} \cmidrule(lr){7-10}
    \textbf{Training data} & \textbf{Length} & \multicolumn{1}{c}{\footnotesize MMLB-D} & \multicolumn{1}{c}{\footnotesize LD-URL} & \multicolumn{1}{c}{\footnotesize SLIDE} & \multicolumn{1}{c|}{\footnotesize AVG.} & \multicolumn{1}{c}{\footnotesize MMLB-D} & \multicolumn{1}{c}{\footnotesize LD-URL} & \multicolumn{1}{c}{\footnotesize SLIDE} & \multicolumn{1}{c|}{\footnotesize AVG.} & \\
    \midrule
    \multirow{2}{*}{\extractsingle} & pool-native & 33.85 & 59.73 & 77.00 & 56.86 & 30.89 & 55.69 & 77.00 & 54.53 & 55.69\,{\scriptsize\makebox[2em][l]{\posdiff{1.3}}} \\
 & long-biased & 31.47 & 55.88 & 75.00 & 54.12 & 29.48 & 59.40 & 75.00 & 54.63 & 54.37\,{\scriptsize\makebox[2em][l]{}} \\
    \midrule
    \multirow{2}{*}{\extractmulti} & pool-native & 32.75 & 64.32 & 77.00 & 58.02 & 31.50 & 54.82 & 81.00 & 55.77 & 56.90\,{\scriptsize\makebox[2em][l]{\posdiff{0.1}}} \\
     & long-biased & 36.29 & 56.75 & 80.00 & 57.68 & 30.72 & 55.79 & 81.00 & 55.84 & 56.76\,{\scriptsize\makebox[2em][l]{}} \\
    \midrule
    \multirow{2}{*}{\reasoningtask} & pool-native & 32.67 & 60.34 & 79.00 & 57.33 & 29.23 & 61.61 & 76.00 & 55.62 & 56.47\,{\scriptsize\makebox[2em][l]{\posdiff{1.7}}} \\
     & long-biased & 32.00 & 55.79 & 79.00 & 55.60 & 28.74 & 58.22 & 75.00 & 53.98 & 54.79\,{\scriptsize\makebox[2em][l]{}} \\
    \bottomrule
    \end{tabular}
    }
\end{table}

\subsection{Short-Data Mixing for Long-Document VQA}
\label{app:complementary_experiment_results:short_mix_long_vqa}
We provide the full per-dataset long-document VQA results for the short-data mixing ablation in \cref{tab:short_mix_long_vqa}, complementing the AVG-only summary plotted in \cref{fig:short_mix_long_vqa_avg}.

\begin{table}[H]
    \centering
    \caption{Short-context data test. We mix in $0\%$ to $80\%$ short-context data during LongPT under the fixed $5B$-token budget. The long-context data uses the 8:2 extraction-to-reasoning long-context mixture. \textbf{Short Data} means the proportion of short-context data. $0\%$ means only using long-context data.}
    \label{tab:short_mix_long_vqa}
    \setlength{\tabcolsep}{4pt}
    \resizebox{0.85\textwidth}{!}{%
    \begin{tabular}{l*{4}{C{1.2cm}}|*{4}{C{1.2cm}}|C{1.2cm}}
    \toprule
    \multirow{2}{*}{\makecell{\textbf{Short}\\\textbf{Data}}} & \multicolumn{4}{c|}{\textbf{64K MMLongBench}} & \multicolumn{4}{c|}{\textbf{128K MMLongBench}} & \multirow{2}{*}{\textbf{AVG.}} \\
    \cmidrule(lr){2-5} \cmidrule(lr){6-9}
    & \multicolumn{1}{c}{\footnotesize MMLB-D} & \multicolumn{1}{c}{\footnotesize LD-URL} & \multicolumn{1}{c}{\footnotesize SLIDE} & \multicolumn{1}{c|}{\footnotesize AVG.} & \multicolumn{1}{c}{\footnotesize MMLB-D} & \multicolumn{1}{c}{\footnotesize LD-URL} & \multicolumn{1}{c}{\footnotesize SLIDE} & \multicolumn{1}{c|}{\footnotesize AVG.} & \\
    \midrule
    0\% & 36.00 & 62.69 & 80.00 & \textbf{59.56} & 34.19 & 56.33 & 77.00 & 55.84 & \textbf{57.70} \\
    20\% & 32.33 & 57.61 & 78.00 & 55.98 & 31.49 & 56.00 & 78.00 & 55.16 & 55.57 \\
    40\% & 34.90 & 56.65 & 80.00 & 57.19 & 30.19 & 61.30 & 79.00 & \textbf{56.83} & 57.01 \\
    60\% & 33.62 & 57.53 & 81.00 & 57.38 & 31.93 & 58.61 & 79.00 & 56.52 & 56.95 \\
    80\% & 33.90 & 57.60 & 80.00 & 57.17 & 31.93 & 55.17 & 81.00 & 56.03 & 56.60 \\
    \bottomrule
\end{tabular}
    }
\end{table}

\subsection{mRoPE Base Frequency}
\label{app:complementary_experiment_results:mrope_base}
Existing studies~\citep{xiong2024effective,blocntkaware,blocntkparts,pengyarn} have shown that increasing the RoPE~\citep{su2024roformer} frequency base during long-context continued pre-training or inference can improve long-context performance.
Dynamic-NTK~\citep{dynamicntk} suggests scaling the frequency base by $t^{\frac{d}{d-2}}$, where $t$ denotes the context-window expansion factor and $d$ is the attention head dimension.

However, LVLMs often use more structured positional encodings than plain 1-D RoPE.
In our experiments, Qwen2.5-VL-7B adopts mRoPE~\citep{wang2024qwen2}, which decomposes rotary embeddings into temporal, height, and width components.
As a result, visual position indices grow more slowly than a flattened 1-D sequence.
It is therefore unclear whether the RoPE-scaling heuristic from Dynamic-NTK, originally developed for LLMs, directly applies to LVLMs with mRoPE.

We conduct an ablation study over the mRoPE frequency base for LVLM long-context training.
By default, we follow Dynamic-NTK and scale the frequency base of the trained model from $1\times 10^6$ to $4\times 10^6$ when extending the context window from 32K to 128K tokens.
We also evaluate two alternative bases, $2\times 10^6$ and $8\times 10^6$, to examine whether decreasing or increasing the default $4\times 10^6$ base improves LongPT performance.

The evaluation results on long-document VQA tasks are shown in \cref{tab:rope_base_vqa}.
Overall, moderately increasing the mRoPE base improves or maintains long-context performance compared with using a smaller base.
Across the evaluated tasks, $2\times 10^6$ and the Dynamic-NTK-scaled base $4\times 10^6$ achieve comparable overall performance, with $4\times 10^6$ slightly outperforming $2\times 10^6$ on \extractmulti and \reasoningtask.
Further increasing the base to $8\times 10^6$ improves some individual metrics, but does not yield consistent gains across tasks and can degrade performance on \extractmulti and \reasoningtask.
These results suggest that moderate mRoPE-base scaling is sufficient for extending LVLMs to longer contexts, while overly aggressive scaling is unnecessary.
Based on this observation and to maintain consistency with the Dynamic-NTK heuristic, we set $4\times 10^6$ as the mRoPE base in our main experiments.

\begin{table}[H]
    \centering
    \caption{Ablation on the mRoPE frequency base for 128K LongPT.
    We grid search the mRoPE base in $\{2\times 10^6, 4\times 10^6, 8\times 10^6\}$, starting from the original base of $1\times 10^6$ used for the 32K context.
    The Dynamic-NTK heuristic~\citep{dynamicntk} suggests $4\times 10^6$ as the default base when extending the context window from 32K to 128K.
    Overall, $2\times 10^6$ and $4\times 10^6$ achieve comparable performance, while further increasing the base to $8\times 10^6$ does not yield consistent gains across tasks.
    }
    \label{tab:rope_base_vqa}
    \setlength{\tabcolsep}{4pt}
    \resizebox{0.95\textwidth}{!}{%
    \begin{tabular}{ll*{4}{C{1.2cm}}|*{4}{C{1.2cm}}|C{1.2cm}}
    \toprule
    & & \multicolumn{4}{c|}{\textbf{64K MMLongBench}} & \multicolumn{4}{c|}{\textbf{128K MMLongBench}} & \multirow{2}{*}{\textbf{AVG.}} \\
     \cmidrule(lr){3-6} \cmidrule(lr){7-10}
    \textbf{Training data} & \textbf{Freq.} & \multicolumn{1}{c}{\footnotesize MMLB-D} & \multicolumn{1}{c}{\footnotesize LD-URL} & \multicolumn{1}{c}{\footnotesize SLIDE} & \multicolumn{1}{c|}{\footnotesize AVG.} & \multicolumn{1}{c}{\footnotesize MMLB-D} & \multicolumn{1}{c}{\footnotesize LD-URL} & \multicolumn{1}{c}{\footnotesize SLIDE} & \multicolumn{1}{c|}{\footnotesize AVG.} & \\
    \midrule
    \multirow{3}{*}{\extractsingle} & $2\times 10^6$ & \textbf{35.52} & 57.07 & 78.00 & \textbf{56.86} & 28.87 & \textbf{58.65} & 78.00 & 55.17 & \textbf{56.02} \\
     & $4\times 10^6$ & 33.85 & \textbf{59.73} & 77.00 & 56.86 & \textbf{30.89} & 55.69 & 77.00 & 54.53 & 55.69 \\
     & $8\times 10^6$ & 33.63 & 52.73 & \textbf{79.00} & 55.12 & 30.83 & 58.10 & \textbf{79.00} & \textbf{55.98} & 55.55 \\
    \midrule
    \multirow{3}{*}{\extractmulti} & $2\times 10^6$ & 31.86 & 63.01 & 76.00 & 56.95 & \textbf{31.60} & 52.31 & 79.00 & 54.30 & 55.63 \\
     & $4\times 10^6$ & 32.75 & \textbf{64.32} & 77.00 & \textbf{58.02} & 31.50 & \textbf{54.82} & \textbf{81.00} & \textbf{55.77} & \textbf{56.90} \\
     & $8\times 10^6$ & \textbf{33.36} & 58.91 & \textbf{79.00} & 57.09 & 29.25 & 54.00 & \textbf{81.00} & 54.75 & 55.92 \\
    \midrule
    \multirow{3}{*}{\reasoningtask} & $2\times 10^6$ & 33.04 & 57.20 & 77.00 & 55.75 & 33.44 & 58.66 & \textbf{77.00} & \textbf{56.36} & 56.05 \\
     & $4\times 10^6$ & 32.67 & \textbf{60.34} & \textbf{79.00} & \textbf{57.33} & 29.23 & \textbf{61.61} & 76.00 & 55.62 & \textbf{56.47} \\
     & $8\times 10^6$ & \textbf{34.33} & 57.86 & 73.00 & 55.07 & \textbf{34.26} & 54.61 & 72.00 & 53.62 & 54.34 \\
    \bottomrule
    \end{tabular}
    }
\end{table}

\subsection{MM-NIAH Generalization}
\label{app:complementary_experiment_results:generalization_niah}
We provide the full 64K and 128K MM-NIAH results in \cref{tab:generalization_niah}, complementing the averaged main-text summary in \cref{fig:generalization_niah}.

\begin{table}[H]
\centering
\caption{\textbf{MM-NIAH generalization.} Ret., Count, and Reas. average text and image needles.}
\label{tab:generalization_niah}
\scriptsize
\captionsetup[subtable]{font=footnotesize,labelformat=parens,justification=centering,skip=1pt}

\begin{subtable}{0.48\textwidth}
    \centering
    \caption{\textbf{64K MM-NIAH}}

    \begin{tabular*}{\linewidth}{@{\extracolsep{\fill}}lcccc@{}}
    \toprule
    \textbf{Model} & Ret. & Count & Reas. & AVG. \\
    \midrule
    \modelname & \textbf{74.83} & \textbf{27.67} & \textbf{67.33} & \textbf{56.61} \\
    Qwen2.5-VL-7B & 50.00 & 6.00 & 27.50 & 27.83 \\
    \midrule
    InternVL3-8B & 67.67 & 14.33 & 59.33 & 47.11 \\
    InternVL3.5-8B & 56.12 & 4.45 & 57.23 & 39.26 \\
    Gemma3-4B & 43.25 & 4.33 & 31.50 & 26.36 \\
    Gemma3-12B & 65.58 & 27.00 & 42.25 & 44.94 \\
    \bottomrule
    \end{tabular*}
\end{subtable}
\hfill
\begin{subtable}{0.48\textwidth}
    \centering
    \caption{\textbf{128K MM-NIAH}}

    \begin{tabular*}{\linewidth}{@{\extracolsep{\fill}}lcccc@{}}
    \toprule
    \textbf{Model} & Ret. & Count & Reas. & AVG. \\
    \midrule
    \modelname & \textbf{57.83} & 8.67 & \textbf{60.33} & \textbf{42.28} \\
    Qwen2.5-VL-7B & 11.33 & 16.33 & 8.83 & 12.17 \\
    \midrule
    InternVL3-8B & 52.33 & 7.33 & 50.00 & 36.56 \\
    InternVL3.5-8B & 3.45 & 0.00 & 2.27 & 1.91 \\
    Gemma3-4B & 29.83 & 1.83 & 28.75 & 20.14 \\
    Gemma3-12B & 48.83 & \textbf{19.67} & 33.75 & 34.08 \\
    \bottomrule
    \end{tabular*}
\end{subtable}
\end{table}

\subsection{VTCBench Generalization}
\label{app:complementary_experiment_results:generalization_vtc}
We provide the full VTCBench-Wild results in \cref{tab:generalization_vtc}. We report retrieval, reasoning, memory, and the sample-count weighted overall score following VTCBench-Wild.

\begin{table}[H]
    \centering
    \caption{VTCBench generalization. We report retrieval, reasoning, memory, and the sample-count weighted overall score following VTCBench-Wild.}
    \label{tab:generalization_vtc}
    \scriptsize
    \begin{tabular*}{0.7\linewidth}{@{\extracolsep{\fill}}lcccc@{}}
    \toprule
    \textbf{Model} & Ret. & Reas. & Mem. & AVG. \\
    \midrule
    \modelname & \textbf{91.75} & \textbf{22.88} & \textbf{40.50} & \textbf{52.73} \\
    Qwen2.5-VL-7B & 91.63 & 15.63 & 33.83 & 48.23 \\
    \midrule
    Qwen3-VL-8B & 89.00 & 11.50 & 33.67 & 45.73 \\
    InternVL3.5-8B & 51.38 & 7.63 & 17.33 & 26.18 \\
    InternVL3.5-38B & 45.81 & 8.75 & 22.33 & 25.93 \\
    Gemma3-27B & 49.38 & 3.75 & 17.33 & 24.05 \\
    \bottomrule
    \end{tabular*}
\end{table}

\subsection{Long-Video Generalization}
\label{app:complementary_experiment_results:generalization_video}
We provide the numeric long-video generalization results in \cref{tab:generalization_video}, complementing the main-text summary in \cref{fig:generalization_video}.

\begin{table}[H]
\centering
\caption{\textbf{Long-video generalization.} We report aggregate scores on Video-MME, MLVU, and LongVideoBench.}
\label{tab:generalization_video}
\scriptsize
\begin{tabular*}{0.7\linewidth}{@{\extracolsep{\fill}}lccc@{}}
\toprule
\textbf{Model} & Video-MME & MLVU & LongVideoBench \\
\midrule
Qwen2.5-VL-7B & 65.1 & 70.2 & 60.43 \\
\modelname & \textbf{67.78} & \textbf{73.55} & \textbf{62.08} \\
\bottomrule
\end{tabular*}
\end{table}

\begin{table}[H]
    \centering
    \caption{Long-document VQA transfer across backbone models. We apply LongPT variants to Qwen3-VL-8B and report results on LongDocURL and SlideVQA. Since Qwen3-VL is already a 256K-context model trained with large-scale long-context pre-training, SFT, and RL, we focus on these two held-out document benchmarks.}
    \label{tab:qwen3_backbone_generalization}
    \setlength{\tabcolsep}{4pt}
    \resizebox{0.9\textwidth}{!}{%
    \begin{tabular}{l*{3}{C{1.2cm}}|*{3}{C{1.2cm}}|C{1.4cm}}
    \toprule
    & \multicolumn{3}{c|}{\textbf{64K MMLongBench}} & \multicolumn{3}{c|}{\textbf{128K MMLongBench}} & \multirow{2}{*}{\textbf{AVG.}} \\
    \cmidrule(lr){2-4} \cmidrule(lr){5-7}
    \textbf{Model} & \multicolumn{1}{c}{\footnotesize LD-URL} & \multicolumn{1}{c}{\footnotesize SLIDE} & \multicolumn{1}{c|}{\footnotesize AVG.} & \multicolumn{1}{c}{\footnotesize LD-URL} & \multicolumn{1}{c}{\footnotesize SLIDE} & \multicolumn{1}{c|}{\footnotesize AVG.} & \\
    \midrule
    Qwen3-VL-8B & 56.36 & 74.00 & 65.18 & 60.11 & 72.00 & 66.05 & 65.62\,{\scriptsize\makebox[2em][l]{}} \\
    + \modelname Recipe & 62.22 & 73.00 & 67.61 & 62.81 & 72.00 & 67.41 & 67.51\,{\scriptsize\makebox[2em][l]{\posdiff{1.9}}} \\
    \bottomrule
    \end{tabular}
    }
\end{table}

\subsection{Generalization across Backbones}
\label{app:complementary_experiment_results:backbone}

We further apply our LongPT recipe to Qwen3-VL-8B~\citep{bai2025qwen3} to test whether the training recipe transfers beyond Qwen2.5-VL. 
However, the Qwen3-VL series already includes 256K-context models trained with 100B tokens in long-context continued pre-training, and Qwen3-VL has undergone additional SFT and RL optimization for long-document tasks~\citep{bai2025qwen3}. 
Therefore, this experiment is not intended as a strict study on context window extension. 
Instead, we use it as a diagnostic to examine whether the observed behavior transfers across backbones. 
We report the performance on long-document VQA tasks and MM-NIAH in \cref{tab:qwen3_backbone_generalization,tab:qwen3_backbone_mmniah}, respectively.

The results show that the final \modelname recipe remains effective even on this stronger long-context backbone.
On long-document VQA, \modelname improves the average score from 65.62 to 67.51.
The gains are more pronounced on MM-NIAH, where the average score increases from 50.03 to 61.75, with consistent improvements across retrieval, counting, and reasoning at both context lengths.
Together with the diagnostic nature of this experiment, these results suggest that the proposed training recipe is not specific to Qwen2.5-VL-7B, but may also improve the long-context behavior of a newer backbone that already incorporates native long-context training.

\begin{table}[t]
    \centering
    \caption{MM-NIAH transfer across backbone models. We report Retrieval, Counting, Reasoning, and average scores at 64K and 128K context lengths. Our recipe achieves better overall MM-NIAH score on Qwen3-VL-8B.}
    \label{tab:qwen3_backbone_mmniah}
    \setlength{\tabcolsep}{3.5pt}
    \resizebox{\textwidth}{!}{%
    \begin{tabular}{l*{4}{C{1.05cm}}|*{4}{C{1.05cm}}|C{1.25cm}}
    \toprule
    & \multicolumn{4}{c|}{\textbf{64K MM-NIAH}} & \multicolumn{4}{c|}{\textbf{128K MM-NIAH}} & \multirow{2}{*}{\textbf{AVG.}} \\
    \cmidrule(lr){2-5} \cmidrule(lr){6-9}
    \textbf{Model} & \multicolumn{1}{c}{\footnotesize Ret.} & \multicolumn{1}{c}{\footnotesize Count} & \multicolumn{1}{c}{\footnotesize Reas.} & \multicolumn{1}{c|}{\footnotesize AVG.} & \multicolumn{1}{c}{\footnotesize Ret.} & \multicolumn{1}{c}{\footnotesize Count} & \multicolumn{1}{c}{\footnotesize Reas.} & \multicolumn{1}{c|}{\footnotesize AVG.} & \\
    \midrule
    Qwen3-VL-8B & 77.83 & 25.33 & 62.33 & 55.17 & 61.67 & 17.00 & 56.00 & 44.89 & 50.03\,{\scriptsize\makebox[2em][l]{}} \\
    + \modelname Recipe & \textbf{83.00} & \textbf{45.67} & \textbf{70.83} & \textbf{66.50} & \textbf{74.17} & \textbf{35.00} & \textbf{61.83} & \textbf{57.00} & \textbf{61.75}\,{\scriptsize\makebox[2em][l]{\posdiff{11.7}}} \\
    \bottomrule
    \end{tabular}
    }
\end{table}

\section{Limitations}
\label{app:limitations}
For the training scale, our systematic study is primarily conducted on 7B/8B-scale LVLMs. 
This choice enables controlled comparisons across data recipes, context lengths, and training budgets, but it also leaves open how the observed trends scale to substantially larger models.
Extending the same study to 30B or 70B-scale LVLMs, or to even longer context windows, like 512K or 1M, would require significantly higher computational cost, since long-context continued pre-training is expensive both in model size and sequence length.
Although our transfer experiment on Qwen3-VL-8B provides initial evidence that the proposed recipe is not tied to a single backbone, a more comprehensive scaling study across larger model families remains an important direction for future work.

For evaluation, our long-document VQA experiments rely on model-based judging to assess answer correctness.
Compared with lexical-overlap metrics, such judges can better handle semantically equivalent answers and free-form generation, but they also introduce additional API cost.
This cost becomes substantial when evaluating many checkpoints, context lengths, and model variants, which limits the frequency and scale of evaluation during long-context training.
Developing more efficient, reliable, and low-cost evaluation protocols for multimodal long-context models is, therefore, an important direction for future research.

\section{Broader Impact}
\label{app:broader_impact}
Reliable long-context capability is important for deploying LVLMs in real-world scenarios that require understanding and reasoning over large multimodal inputs, such as long documents, webpages, videos, and agentic workflows.
This work studies how to build such capability through long-context continued pre-training, with a particular focus on constructing effective multimodal long-context data under a modest training budget.
Our findings suggest that carefully designed long-document VQA data can provide useful supervision for evidence retrieval and reasoning over long visual-textual contexts, and can transfer beyond documents to broader multimodal long-context tasks.
We hope these results can help the community develop more data-efficient training recipes for LVLMs, reducing the need for excessive token budgets while improving models' ability to use long multimodal context.

\end{document}